\documentclass[journal]{IEEEtran}

\usepackage[utf8]{inputenc}
\usepackage{booktabs,tabularx}
\usepackage{times}
\usepackage{epsfig}
\usepackage{graphicx}
\usepackage{amsmath}
\usepackage{amssymb}
\usepackage{pifont}
\usepackage[table,usenames,dvipsnames,svgnames]{xcolor} 
\usepackage{colortbl}
\usepackage{wrapfig}
\usepackage{tabu}
\usepackage{algorithm}
\usepackage[noend]{algpseudocode}
\usepackage{upgreek}
\usepackage{cite}
\usepackage{afterpage}
\usepackage{pdflscape}

\makeatother

\usepackage{blindtext}

\newcommand{\dashrule}[1][black]{%
	\color{#1}\rule[\dimexpr.5ex-.2pt]{4pt}{.4pt}\xleaders\hbox{\rule{4pt}{0pt}\rule[\dimexpr.5ex-.2pt]{4pt}{.4pt}}\hfill\kern0pt%
}
\newcommand{\rulecolor}[1]{%
	\def\CT@arc@{\color{#1}}%
}

\usepackage{booktabs}
\usepackage{multirow}

\usepackage{stackengine}
\usepackage{scalerel}
\newlength\lthk
\def\bline{\rule{2ex}{\lthk}}
\def\slash{\rotatebox{60}{\bline}}
\def\parallelogram{\stackMath\scalerel*{%
		\def\stackalignment{l}{\stackunder[-.5\lthk]{%
				\def\stackalignment{r}\stackon[-.5\lthk]{\slash\rule{.866ex}{0ex}\slash}{\bline}}%
			{\bline}}}{\square}%
}

\definecolor{verylightgray}{gray}{0.95}

\usepackage[unicode,hyperindex,plainpages=false,pdftex,hidelinks]{hyperref}
\hypersetup{
	colorlinks=true,
	linkcolor=BrickRed,
	citecolor=OliveGreen,
	filecolor=magenta,
	urlcolor=cyan
}

\begin{document}
%
\title{BoxCars: Improving Fine-Grained Recognition\\of Vehicles using 3D Bounding Boxes\\in Traffic Surveillance}
%
%
%

\author{Jakub Sochor, Jakub Špaňhel, Adam Herout
\thanks{The authors are with Brno University of Technology, Faculty of Information Technology, Centre of Excellence IT4Innovations, Czech Republic \{isochor,ispanhel,herout\}@fit.vutbr.cz}
\thanks{Jakub Sochor is a Brno Ph.D. Talent Scholarship Holder --- Funded by the Brno City Municipality.}
\thanks{Digital Object Identifier 10.1109/TITS.2018.2799228}%
\thanks{1524-9050 \copyright\ IEEE}%
}

\markboth{IEEE Transactions on Intelligent Transportation Systems}%
{Sochor \MakeLowercase{\textit{et al.}}: BoxCars: Improving Fine-Grained Recognition of Vehicles using 3D Bounding Boxes in Traffic Surveillance}


\maketitle

\begin{abstract}
In this paper, we focus on fine-grained recognition of vehicles mainly in traffic surveillance applications. We propose an approach that is orthogonal to recent advancements in fine-grained recognition (automatic part discovery, bilinear pooling). Also, in contrast to other methods focused on fine-grained recognition of vehicles, we do not limit ourselves to a frontal/rear viewpoint, but allow the vehicles to be seen from any viewpoint. 
Our approach is based on 3D bounding boxes built around the vehicles. The bounding box can be automatically constructed from traffic surveillance data. For scenarios where it is not possible to use precise construction, we propose a method for an estimation of the 3D bounding box. 
The 3D bounding box is used to normalize the image viewpoint by ``unpacking'' the image into a plane. We also propose to randomly alter the color of the image and add a rectangle with random noise to a random position in the image during the training of Convolutional Neural Networks. 
We have collected a large fine-grained vehicle dataset BoxCars116k, with 116k images of vehicles from various viewpoints taken by numerous surveillance cameras. We performed a number of experiments which show that our proposed method significantly improves CNN classification accuracy (the accuracy is increased by up to 12 percentage points and the error is reduced by up to 50\,\% compared to CNNs without the proposed modifications). We also show that our method outperforms state-of-the-art methods for fine-grained recognition.
\end{abstract}

\begin{figure}[h]
	\includegraphics[width=0.5\linewidth]{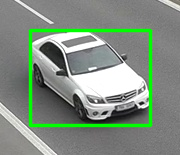}%
	\includegraphics[width=0.5\linewidth]{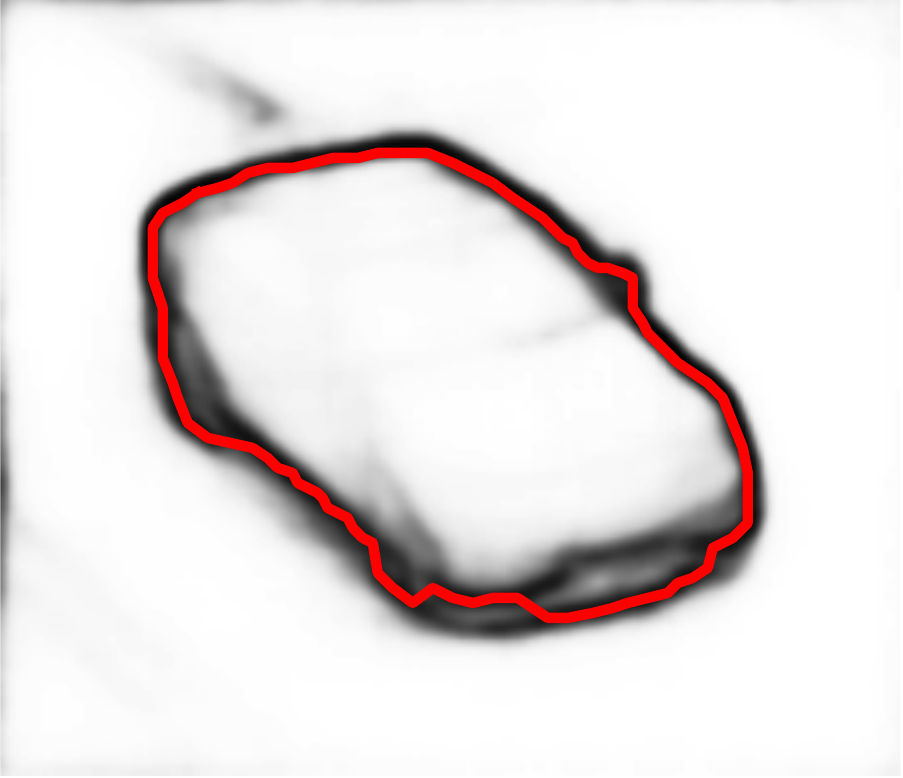}\\
	\includegraphics[width=0.5\linewidth]{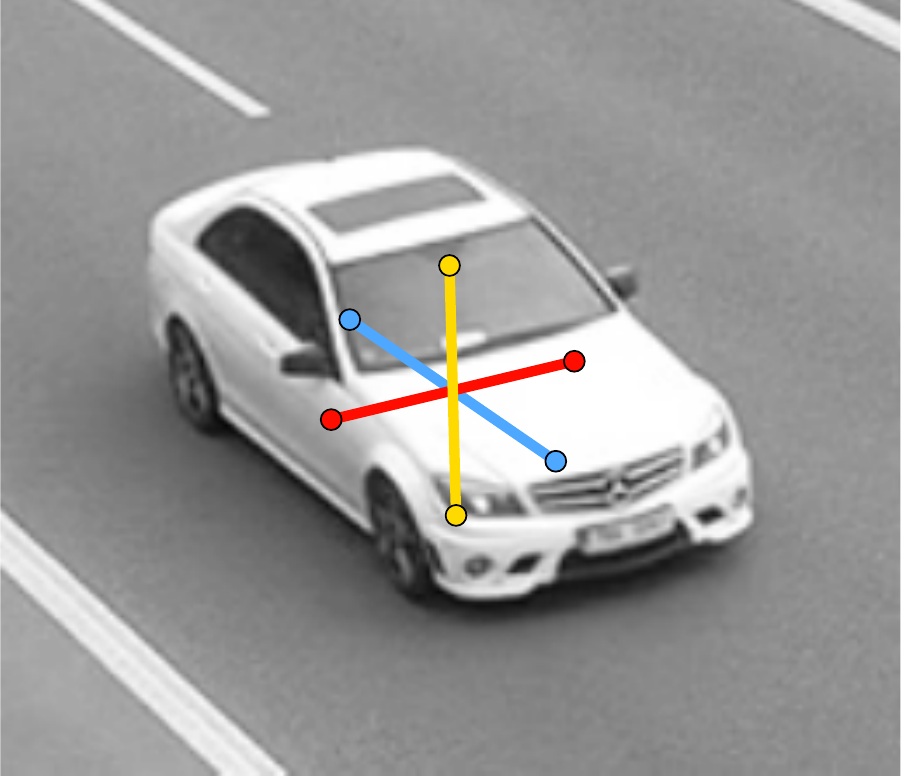}%
	\includegraphics[width=0.5\linewidth]{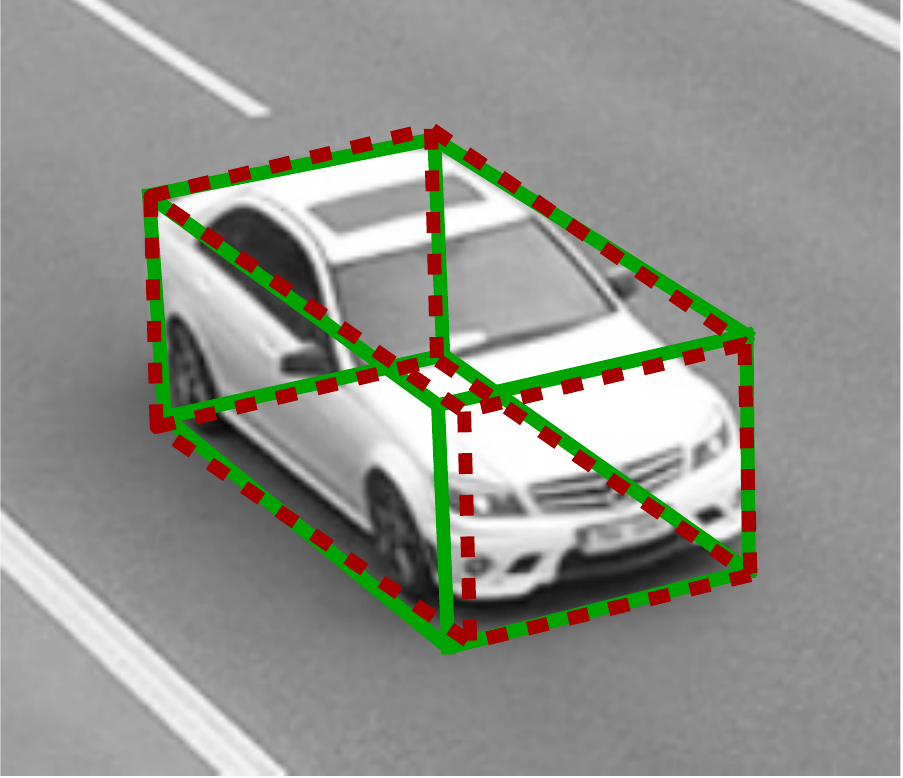}%
	\caption{Example of automatically obtained 3D bounding box used for fine-grained vehicle classification. \textbf{Top left:} vehicle with 2D bounding box annotation, \textbf{top right:} estimated contour, \textbf{bottom left:} estimated directions to vanishing points, \textbf{bottom right:} 3D bounding box automatically obtained from surveillance video (green) and our estimated 3D bounding box (red).} \label{fig:Teaser}
\end{figure}

\section{Introduction} 
\label{sec:Introduction}

Fine-grained recognition of vehicles is interesting, both from the application point of view (surveillance, data retrieval, etc.) and from the point of view of general fine-grained recognition research applicable in other fields. For example, Gebru et al. \cite{Gebru2017} proposed an estimation of demographic statistics based on fine-grained recognition of vehicles.
In this article, we are presenting methodology which considerably increases the performance of multiple state-of-the-art CNN architectures in the task of fine-grained vehicle recognition.
We target the traffic surveillance context, namely images of vehicles taken from an \textbf{arbitrary viewpoint} -- we do not limit ourselves to frontal/rear viewpoints. As the images are obtained from surveillance cameras, they have challenging properties -- they are often small and taken from very general viewpoints (high elevation). We also construct the training and testing sets from images from different cameras as it is common for surveillance applications that it is not known a priori under which viewpoint the camera will be observing the road. 

Methods focused on the fine-grained recognition of vehicles usually have some limitations -- they can be limited to frontal/rear viewpoints or use 3D CAD models of all the vehicles. Both these limitations are rather impractical for large scale deployment. There are also methods for fine-grained recognition in general which were applied on vehicles. The methods recently follow several main directions -- automatic discovery of parts \cite{Krause2015,Simon2015}, bilinear pooling \cite{Lin2015Bilinear,Gao2016}, or exploiting structure of fine-grained labels \cite{Xie2015,Zhou2016}.
Our method is not limited to any particular viewpoint and it does not require 3D models of vehicles at all. 

We propose an orthogonal approach to these methods and use CNNs with a modified input to achieve better image normalization and data augmentation (therefore, our approach can be combined with other methods).  We use 3D bounding boxes around vehicles to normalize vehicle image (see Figure \ref{fig:UnpackVptRastExample} for examples). This work is based on our previous conference paper \cite{Sochor2016BoxCars}; it pushes the performance further and we mainly propose a new method on how to build the 3D bounding box without any prior knowledge (see Figure \ref{fig:Teaser}). Our input modifications are able to significantly increase the classification accuracy (up to \textbf{12 percentage points}, classification error is reduced by up to \textbf{50\,\%}). 

The key contributions of the paper are:
\begin{itemize}
	\item Complex and thorough evaluation of our previous method~\cite{Sochor2016BoxCars}.
	\item Our novel data augmentation techniques further improve the results of the fine-grained recognition of vehicles relative both to our previous method and other state-of-the-art methods (Section \ref{sec:MethodologyDataAug}). 
	\item We remove the requirement of the previous method \cite{Sochor2016BoxCars} to know the 3D bounding box by estimating the bounding box both at training and test time (Section \ref{sec:Methodology3DBBEst}).
	\item We collected more samples to the BoxCars dataset, increasing the dataset size almost twice (Section~\ref{sec:Dataset}).
\end{itemize}

We will make the collected dataset and source codes for the proposed algorithm publicly available\footnote{\url{https://medusa.fit.vutbr.cz/traffic}} for future reference and comparison.
\section{Related Work} \label{sec:SOTA}
In order to provide context to the proposed method, we present a summary of existing fine-grained recognition methods (both general and focused on vehicles). 

\subsection{General Fine-Grained Object Recognition} 



We divide the fine-grained recognition methods from recent literature into several categories as they usually share some common traits.
Methods exploiting annotated model parts (e.g. \cite{Huang2016,Zhang2016_TIP})
are not discussed in detail as it is not common in fine-grained datasets of vehicles to have the parts annotated.

\subsubsection{Automatic Part Discovery}
Parts of classified objects may be discriminatory and provide lots of information for the fine-grained classification task. However, it is not practical to assume that the location of such parts is known a priori as it requires significantly more annotation work. Therefore, several papers \cite{Yang2012,Duan2012,Yao2012,Krause2014,Simon2015,Krause2015,Zhang2016} have dealt with this problem and proposed methods how to automatically (during both training and test time) discover and localize such parts. The methods differ mainly in the ways in which they are used for the discovery of discriminative parts. The features extracted from the parts are  usually classified by SVMs.


\subsubsection{Methods using Bilinear Pooling}
Lin et al. \cite{Lin2015Bilinear} use only convolutional layers from the net for extraction of features which are classified by a bilinear classifier \cite{Pirsiavash2009}. 
Gao et al. \cite{Gao2016} followed the path of bilinear pooling and proposed a method for Compact Bilinear Pooling getting the same accuracy as the full bilinear pooling with a significantly lower number of features. 

\subsubsection{Other Methods}
Xie et al. \cite{Xie2015} proposed to use a hyper-class for data augmentation and regularization of fine-grained deep learning. 
Zhou et al. \cite{Zhou2016} use CNN with Bipartite Graph Labeling to achieve better accuracy by exploiting the fine-grained annotations and coarse body type (e.g. Sedan, SUV).
Lin et al. \cite{Lin2015} use three neural networks for simultaneous localization, alignment and classification of images. Each of these three networks does one of the three tasks and they are connected into one bigger network.
Yao et al. \cite{Yao2012} proposed an approach which uses responses to random templates obtained from  images and classifies merged representations of the response maps by SVM. 
Zhang et al. \cite{Zhang2012} use pose normalization kernels and their responses warped into a feature vector.
Chai et al. \cite{Chai2012} propose to use segmentation for fine-grained recognition to obtain the foreground parts of an image. A similar approach was also proposed by Li et al. \cite{Li2015}; however, the authors use a segmentation algorithm which is optimized and fine-tuned for the purpose of fine-grained recognition. 
Finally, Gavves et al. \cite{Gavves2015} propose to use object proposals to obtain the foreground mask and unsupervised alignment to improve fine-grained classification accuracy.

\subsection{Fine-Grained Recognition of Vehicles} 
The goal of fine-grained recognition of vehicles is to identify the exact type of the vehicle, that is its make, model, submodel, and model year. The recognition system focused only on vehicles (in relation to general fine-grained classification of birds, dogs, etc.) can benefit from that the vehicles are rigid, have some distinguishable landmarks (e.g. license plates), and rigorous mo\-dels (e.g. 3D CAD models) can be available. 

\subsubsection{Methods Limited to Frontal/Rear Images of Vehicles}

There is a multitude of papers \cite{Petrovic2004,Dlagnekov2005,Clady2008,Pearce2011,Psyllos2011,Lee2013,Zhang2013,Llorca2014} using a common approach: they detect the license plate (as a common landmark) on the vehicle and extract features from the area around the license plate as the front/rear parts of vehicles are usually discriminative. 

There are also papers \cite{Zhang2014,Hsieh2014,Hu2015ITS,Liao2015,Baran2015,He2015} directly extracting features from frontal images of vehicles by different methods and optionally exploiting the standard structure of parts on the frontal mask of car (e.g. headlights).

\subsubsection{Methods based on 3D CAD Models}
There were several approaches on how to deal with viewpoint variance using synthetic 3D models of vehicles. Lin et al. \cite{Lin2014} propose to jointly optimize 3D model fitting and fine-grained classification, Hsiao et al. \cite{Hsiao2014} use detected contour and align the 3D model using 3D chamfer matching. Krause et al. \cite{Krause2013}  propose to use synthetic data to train geometry and viewpoint classifiers for the 3D model and 2D image alignment. Prokaj et al. \cite{Prokaj2009} propose to detect SIFT features on the vehicle image and on every 3D model seen from a set of discretized viewpoints.

\subsubsection{Other Methods}
Gu et al. \cite{Gu2013} propose extracting the center of a vehicle and roughly estimate the viewpoint from the bounding box aspect ratio. Then, they use different Active Shape Models for alignment of data taken from different viewpoints and use segmentation for background removal.

 Stark et al. \cite{Stark2012} propose using an extension of Deformable Parts Model (DPM) \cite{Felzenszwalb2010} to be able to handle multi-class recognition. The model is represented by latent linear multi-class SVM with HOG \cite{Dalal2005} features. The authors show that the system outperforms different methods based on Locally-constrained Linear Coding \cite{Wang2010} and HOG. The recognized vehicles are used for eye-level camera calibration.

Liu et al. \cite{Liu2016_DeepRelative} use deep relative distance trained on a vehicle re-identification task  and propose training the neural net with Coupled Clusters Loss instead of triplet loss. 
Boonsim et al.~\cite{Boonsim2016} propose a method for fine-grained recognition of vehicles at night. The authors use relative position and shape of features vi\-si\-ble at night (e.g. lights, license plates) to identify the make\&model of a vehicle, which is visible from the rear side.

Fang et al. \cite{Fang2016} propose using an approach based on detected parts. The parts are obtained in an unsupervised manner as high activations in a mean response across channels of the last convolutional layer of used CNN. The authors in \cite{Hu2017} introduce spatially weighted pooling of convolutional features in CNNs to extract important features from the image. 

\subsubsection{Summary of Existing Methods}
Existing methods for the fine-grained classification of vehicles usually have significant limitations. They are either limited to frontal/rear viewpoints \cite{Petrovic2004,Dlagnekov2005,Clady2008,Pearce2011,Psyllos2011,Lee2013,Zhang2013,Llorca2014,Zhang2014,Hsieh2014,Hu2015ITS,Liao2015,Baran2015,He2015} or require some knowledge about 3D models of the vehicles \cite{Prokaj2009,Krause2013,Hsiao2014,Lin2014} which can be impractical when new models of vehicles emerge. 

Our proposed method does not have such limitations. The method works with arbitrary viewpoints and we require only 3D bounding boxes of vehicles. The 3D bounding boxes can either be automatically constructed from traffic video surveillance data \cite{Dubska2014,Dubska2015ITS} or we propose a method on how to estimate the 3D bounding boxes both at training and test time from single images (see Section \ref{sec:Methodology3DBBEst}).

\subsection{Datasets for Fine-Grained Recognition of Vehicles}
There is a large number of datasets of vehicles (e.g \cite{ILSVRC15,Matzen2013}) 
which are usable mainly for vehicle detection, pose estimation, and other tasks. However, these datasets do not contain annotations of the precise vehicles' make and model.

When it comes to the fine-grained recognition datasets, there are some \cite{Stark2012,Krause2013,Lin2014,Liao2015} which are relatively small in number of samples or classes. Therefore, they are impractical for the training of CNN and deployment of real world traffic surveillance applications.


Yang et al. \cite{Yang2015} published a large dataset \emph{CompCars}. The dataset consists of a web-nature part, made of 136k of vehicles from 1\,600 classes taken from different viewpoints. It also contains a surveillance-nature part with 50k frontal images of vehicles taken from surveillance cameras.

Liu et al. \cite{Liu2016_ReId} published dataset \emph{VeRi\--776} for the vehicle re-identification task. The dataset contains over 50k images of 776 vehicles captured by 20 cameras covering an 1.0 km$^2$ area in 24 hours. Each vehicle is captured by $2\sim18$ cameras under different viewpoints, illuminations, resolutions and occlusions. The dataset also provides various attributes, such as bounding boxes, vehicle types, and colors.

\subsection{Vehicle Detection}
In traffic surveillance applications, it is common that prior fine-grained vehicle classification is necessary to detect vehicles; therefore, we include a brief overview of existing methods for vehicle detection. It is possible to use standard object detectors -- either based on convolutional neural networks \cite{Ren2015,Redmon2015}, AdaBoost \cite{Dollar2014}, Deformable Part Models \cite{Felzenszwalb2010,Felzenszwalb2010PAMI} or Hough Transformation \cite{Gall2011}.
There were also attempts to improve specifically vehicle detection based on geometric information \cite{Wang2017}, during night \cite{Salvi2014}, or to increase the accuracy of localization of occluded vehicles \cite{Soto2017}. 
\section{Proposed Methodology for Fine-Grained Recognition of Vehicles}
\label{sec:Methodology}
 
In agreement with recent progress in the Convolutional Neural Networks \cite{Taigman2014,Krizhevsky2012,Chatfield2014}, we use CNN for both classification and verification (determining whether a pair of vehicles has the same type). However, we propose to use several data normalization and augmentation techniques to significantly boost the classification performance (up to $50\,\%$ error reduction compared to base net). We utilize information about 3D bounding boxes obtained from traffic surveillance camera \cite{Dubska2014}.  Finally, in order to increase the applicability of our method to scenarios where the 3D bounding box is not known, we propose an algorithm for bounding box estimation both at training and test time.

\subsection{Image Normalization by Unpacking the 3D Bounding Box} 
\label{sec:Unpacking}

\begin{figure}
	\centering
	\includegraphics[width=0.25\linewidth]{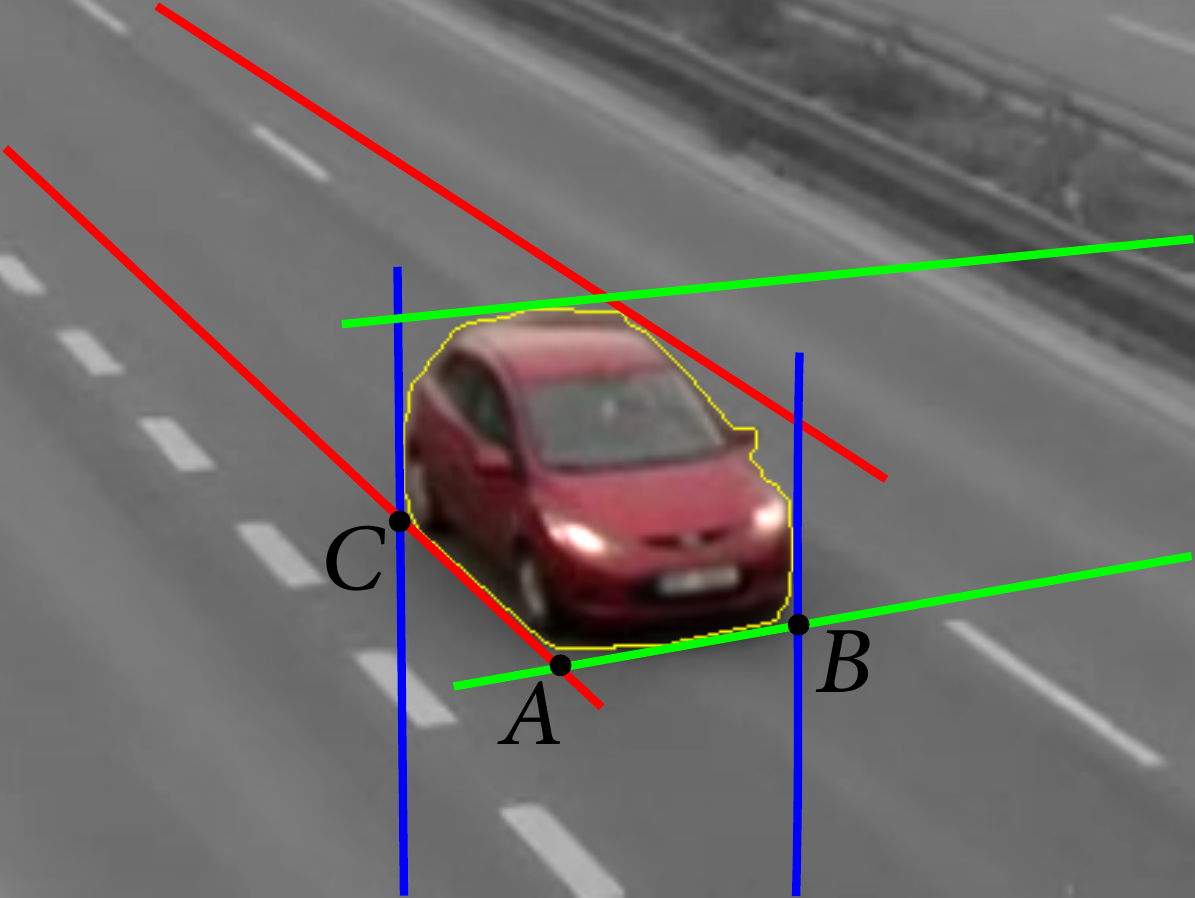}%
	\includegraphics[width=0.25\linewidth]{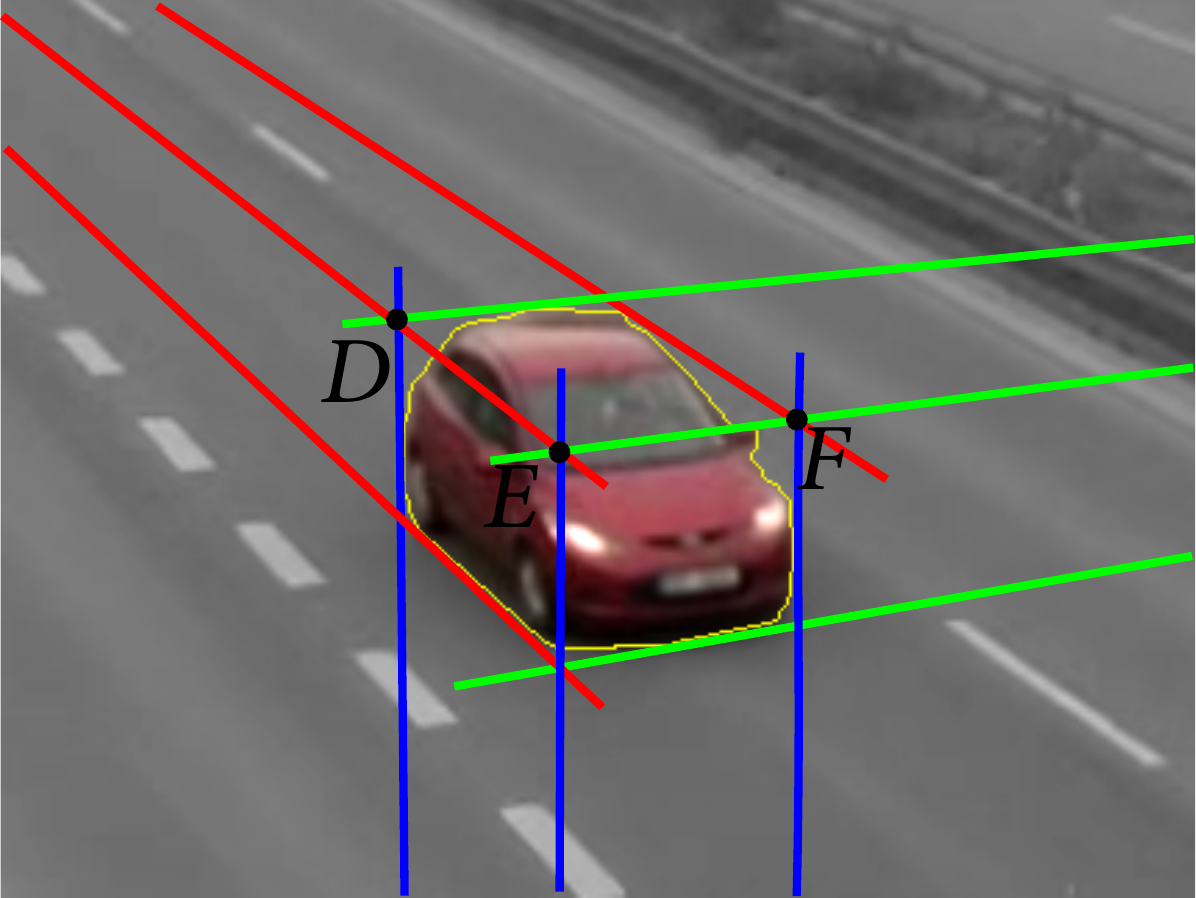}%
	\includegraphics[width=0.25\linewidth]{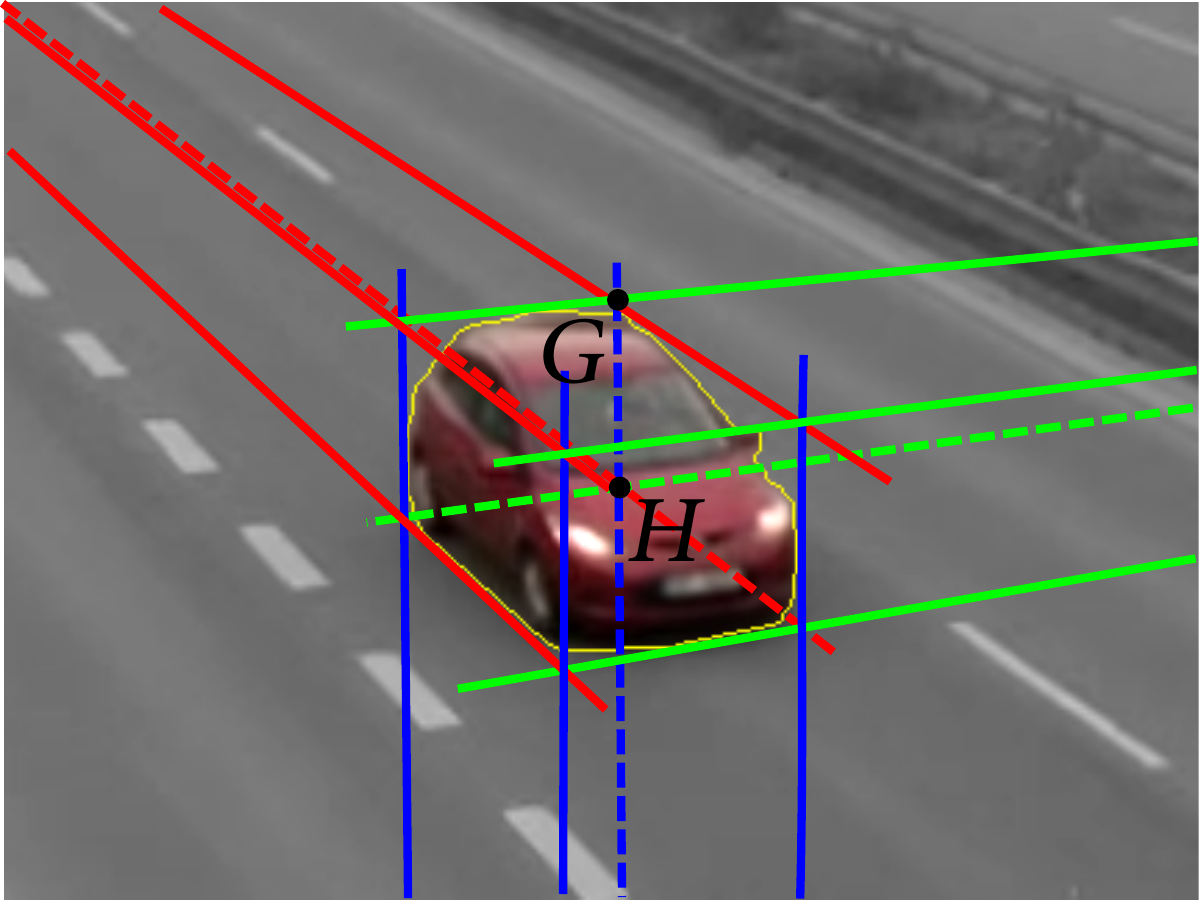}%
	\includegraphics[width=0.25\linewidth]{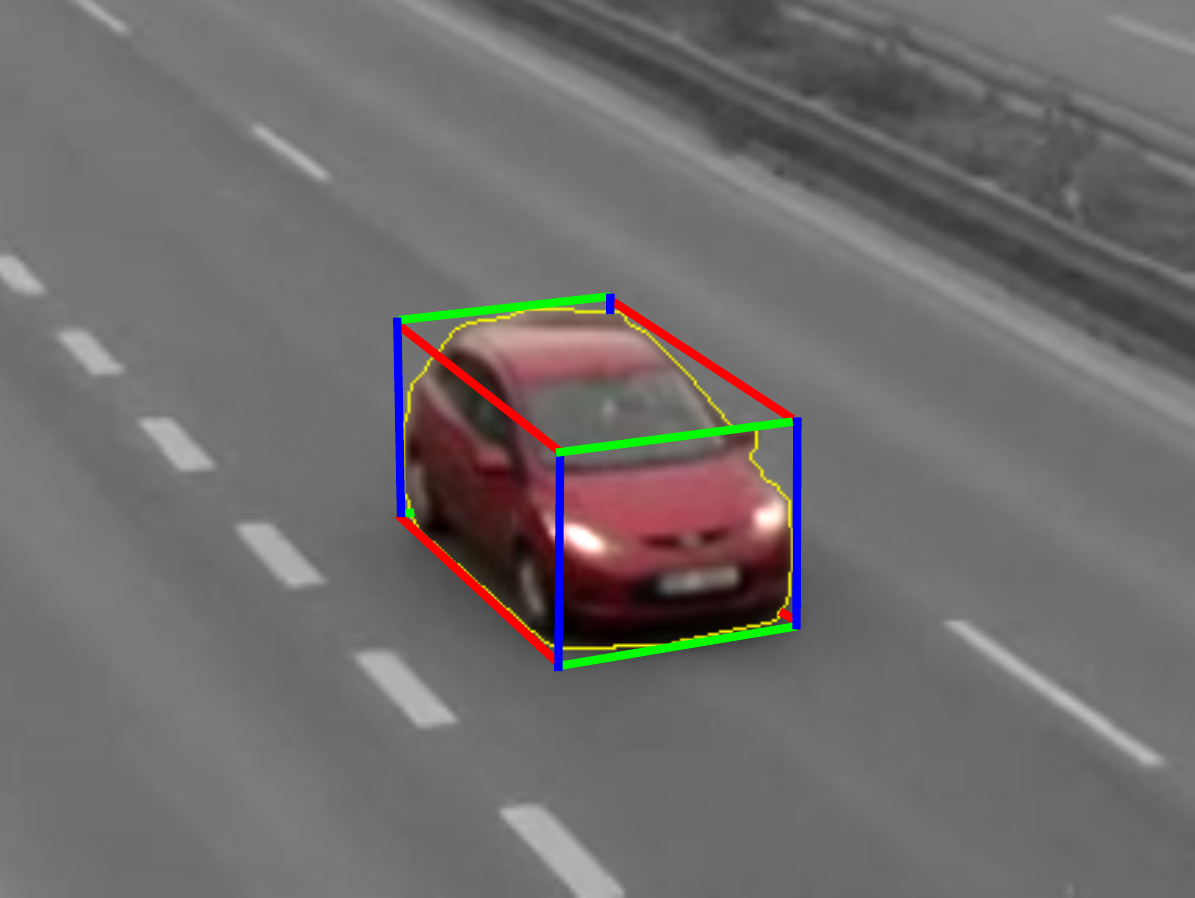}
	\caption{3D bounding box construction process. Each set of lines with the same color intersects in one vanishing point. See the original paper for full details \cite{Dubska2014}. The image was adopted from the paper with the authors' permission.} \label{fig:3DBBConstruction}
\end{figure}

\begin{figure}[t]
	\centering
	\includegraphics[width=0.20\linewidth,height=1.4cm]{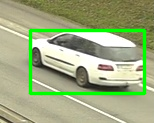}%
	\includegraphics[width=0.20\linewidth,height=1.4cm]{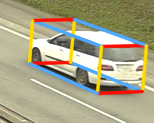}%
	\includegraphics[width=0.20\linewidth,height=1.4cm]{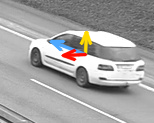}%
	\includegraphics[width=0.20\linewidth,height=1.4cm]{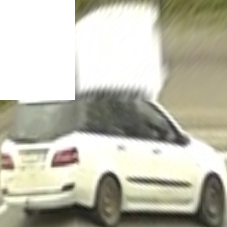}%
	\includegraphics[width=0.20\linewidth,height=1.4cm]{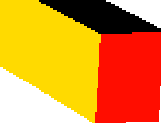}\\
	\includegraphics[width=0.20\linewidth,height=1.4cm]{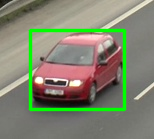}%
	\includegraphics[width=0.20\linewidth,height=1.4cm]{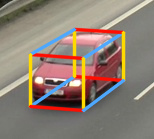}%
	\includegraphics[width=0.20\linewidth,height=1.4cm]{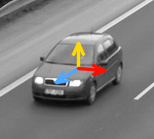}%
	\includegraphics[width=0.20\linewidth,height=1.4cm]{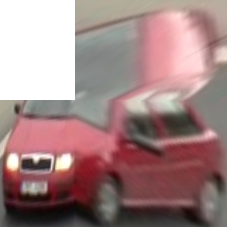}%
	\includegraphics[width=0.20\linewidth,height=1.4cm]{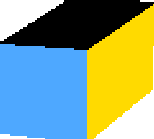}
	\caption{Examples of data normalization and auxiliary data fed to nets. \textbf{Left to right:} vehicle with 2D bounding box, computed 3D bounding box, vectors encoding viewpoints on the vehicle (\textbf{View}), unpacked image of the vehicle (\textbf{Unpack}), and rasterized 3D bounding box fed to the net (\textbf{Rast}). } \label{fig:UnpackVptRastExample}
\end{figure}

We based our work on 3D bounding boxes proposed by \cite{Dubska2014} (Fig.~\ref{fig:UnpackVptRastExample}) which can be automatically obtained for each vehicle seen by a surveillance camera (see Figure \ref{fig:3DBBConstruction} for schematic 3D bounding box construction process or the original paper \cite{Dubska2014} for further details). These boxes allow us to identify the side, roof, and front (or rear) side of vehicles in addition to other information about the vehicles. We use these localized segments to normalize the image of the observed vehicles (considerably boosting the recognition performance).

\begin{figure}[t]
	\centering
	\includegraphics[width=0.8\linewidth]{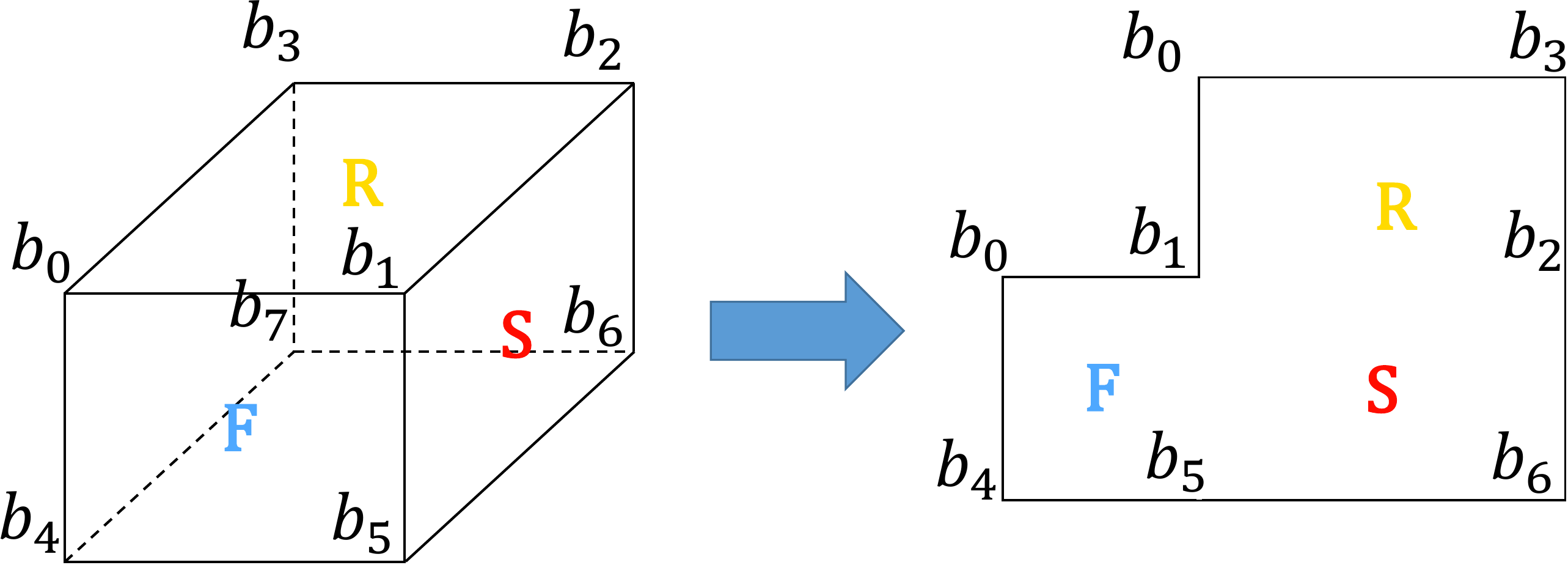}			
	\caption{3D bounding box and its unpacked version.}
	\label{fig:UnpackingSchema}	
\end{figure}

The normalization is done by unpacking the image into a~plane. The plane contains rectified versions of the front/rear ($\mathbf{F}$), side ($\mathbf{S}$), and roof ($\mathbf{R}$). These parts are adjacent to each other (Fig.~\ref{fig:UnpackingSchema}) and they are organized into the final matrix $\mathbf{U}$:
\begin{equation}
\mathbf{U} = \left( \begin{array}{cc}
\mathbf{0} & \mathbf{R}\\
\mathbf{F} & \mathbf{S} \end{array} \right) \label{eq:UnpackMatrix}
\end{equation}

The unpacking itself is done by obtaining homography between points $b_i$ (Fig. \ref{fig:UnpackingSchema}) and perspective warping parts of the original image. The left top submatrix is filled with zeros. This unpacked version of the vehicle is used instead of the original image to feed the net. The unpacking is beneficial as it localizes parts of the vehicles, normalizes their position in the image and it does all that without the necessity of using DPM or other algorithms for part localization. 
Later in the text, we will refer to this normalization method as \textbf{Unpack}.

\subsection{Extended Input to the Neural Nets}

\begin{figure}[t!]
	\centering
	\includegraphics[width=0.25\linewidth]{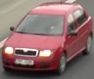}%
	\includegraphics[width=0.25\linewidth]{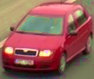}%
	\includegraphics[width=0.25\linewidth]{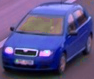}%
	\includegraphics[width=0.25\linewidth]{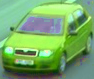}\\
	\includegraphics[width=0.25\linewidth]{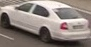}%
	\includegraphics[width=0.25\linewidth]{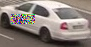}%
	\includegraphics[width=0.25\linewidth]{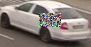}%
	\includegraphics[width=0.25\linewidth]{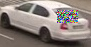}%
	\caption{Examples of proposed data augmentation techniques. Left most image contains the original cropped image of the vehicle and other images contains augmented versions of the image (\textbf{Top -- Color}, \textbf{Bottom -- ImageDrop}). } \label{fig:ImgDropAlterHSVExample}
\end{figure}

It it possible to infer additional information about the vehicle from the 3D bounding box and we found out that these data slightly improve the classification and verification performance.  
One piece of this auxiliary information is the encoded viewpoint (direction from which the vehicle is observed). 
We also add a rasterized 3D bounding box as an additional input to the CNNs.  Compared to our previously proposed auxiliary data fed to the net \cite{Sochor2016BoxCars}, we handle frontal and rear vehicle sides differently. 

\textbf{View.}
The viewpoint is extracted from the orientation of the 3D bounding box -- Fig.~\ref{fig:UnpackVptRastExample}.
We encode the viewpoint as three 2D vectors $v_i$, where $i \in \left\lbrace f, s, r \right\rbrace$ (\emph{front/rear}, \emph{side}, \emph{roof}) and pass them to the net. Vectors $v_i$ are connecting the center of the bounding box with the centers of the box's faces. Therefore, it can be computed as $v_i = \overrightarrow{C_cC_i}$. Point $C_c$ is the center of the bounding box and it can be obtained as 
the intersection of diagonals $\overleftrightarrow{b_2b_4}$ and $\overleftrightarrow{b_5b_3}$.
Points $C_i$ for $i \in \left\lbrace f, s, r \right\rbrace$ denote the centers of each face, again computed as intersections of face diagonals. In contrast to our previous approach \cite{Sochor2016BoxCars}, which did not take the direction of the vehicle into account; instead, we encode the information about the vehicle direction ($d = 1$ for vehicles going to camera, $d=0$ for vehicles going from the camera), in order to determine which side of the bounding box is the frontal one. 
The vectors are normalized to have a unit size; storing them with a different normalization (e.g. the front one normalized, the other in the proper ratio) did not improve the results.

\textbf{Rast.}
Another way of encoding the viewpoint and also the relative dimensions of vehicles is to rasterize the 3D bounding box and use it as an additional input to the net. The rasterization is done separately for all sides, each filled by one color. The final rasterized bounding box is then a four-channel image containing each visible face rasterized in a different channel. Formally, point $p$ of the rasterized bounding box $\mathbf{T}$ is obtained as 
\begin{equation}
\mathbf{T}_{p} = \left\lbrace \begin{array}{cc}
(1,0, 0, 0) & p \in \parallelogram b_0b_1b_4b_5\ \mathrm{and}\ $d = 1$\\
(0,1,0,0) & p \in \parallelogram b_0b_1b_4b_5\ \mathrm{and}\ $d = 0$\\
(0,0,1,0) & p \in \parallelogram b_1b_2b_5b_6\\
(0,0,0,1) & p \in \parallelogram b_0b_1b_2b_3\\
(0,0, 0,0) & \mbox{otherwise}
\end{array} \right. \label{eq:Rasterization}
\end{equation}
where $\parallelogram b_0b_1b_4b_5$ denotes the quadrilateral defined by points $b_0$, $b_1$, $b_4$ and $b_5$ in Figure~\ref{fig:UnpackingSchema}.

Finally, the 3D rasterized bounding box is cropped by the 2D bounding box of the vehicle. For an example, see Figure~\ref{fig:UnpackVptRastExample}, showing rasterized bounding boxes for different vehicles taken from different viewpoints.

\subsection{Additional Training Data Augmentation}
\label{sec:MethodologyDataAug}

In order to increase the diversity of the training data, we propose additional data augmentation techniques. The first one (denoted as \textbf{Color}) deals with the fact that for fine-grained recognition of vehicles (and some other objects), their color is irrelevant. The other method (\textbf{Image\-Drop}) deals with some potentially missing parts of the vehicle. Examples of the data augmentation are shown in Figure~\ref{fig:ImgDropAlterHSVExample}. Both these augmentation techniques are done only with predefined probability during training, otherwise they are not modified. During testing, we do not modify the images at all.

The results presented in Section~\ref{sec:SingleModificationImprovementExperiments} show that both these modifications improve the classification accuracy both in combination with other presented techniques or by themselves.

\textbf{Color.} In order to increase training samples color va\-ri\-abi\-li\-ty, we propose to randomly alternate the color of the image. The alternation is done in the HSV color space by adding the same random values to each pixel in the image (each HSV channel is processed separately). 

\textbf{ImageDrop.} Inspired by Zeiler et al. \cite{Zeiler2014}, who evaluated the influence of covering a part of the input image on the probability of the ground truth class, we take this a step further and in order to deal with missing parts on the vehicles, we take a~random rectangle in the image and fill it with random noise, effectively dropping any information contained in that part of the image.

\subsection{Estimation of 3D Bounding Box from a Single Image} \label{sec:Methodology3DBBEst}
As the results (Section \ref{sec:Experiments}) show, the most important part of the proposed algorithm is \textbf{Unpack} followed by \textbf{Color} and \textbf{ImageDrop}. However, the 3D bounding box is required for unpacking the vehicles and we acknowledge that there may be scenarios when such information is not available. For these cases, we propose a method on how to estimate the 3D bounding box for both training and test time when only limited information is available. 

\begin{figure}[t!]
	\centering
	\includegraphics[width=0.2\linewidth]{./images/22979_2/2DBB}%
	\includegraphics[width=0.2\linewidth]{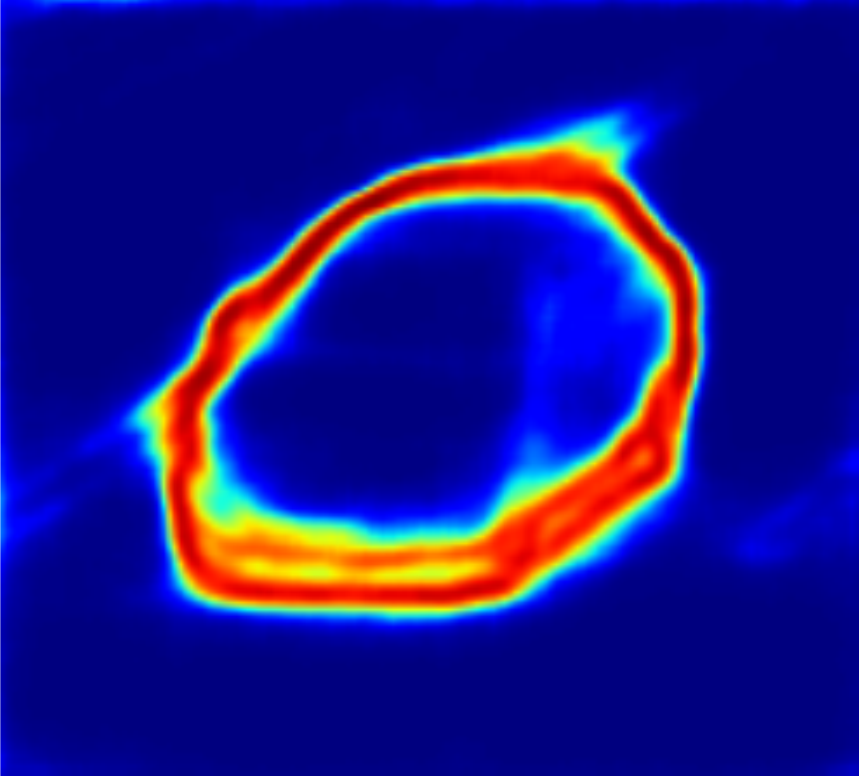}%
	\includegraphics[width=0.2\linewidth]{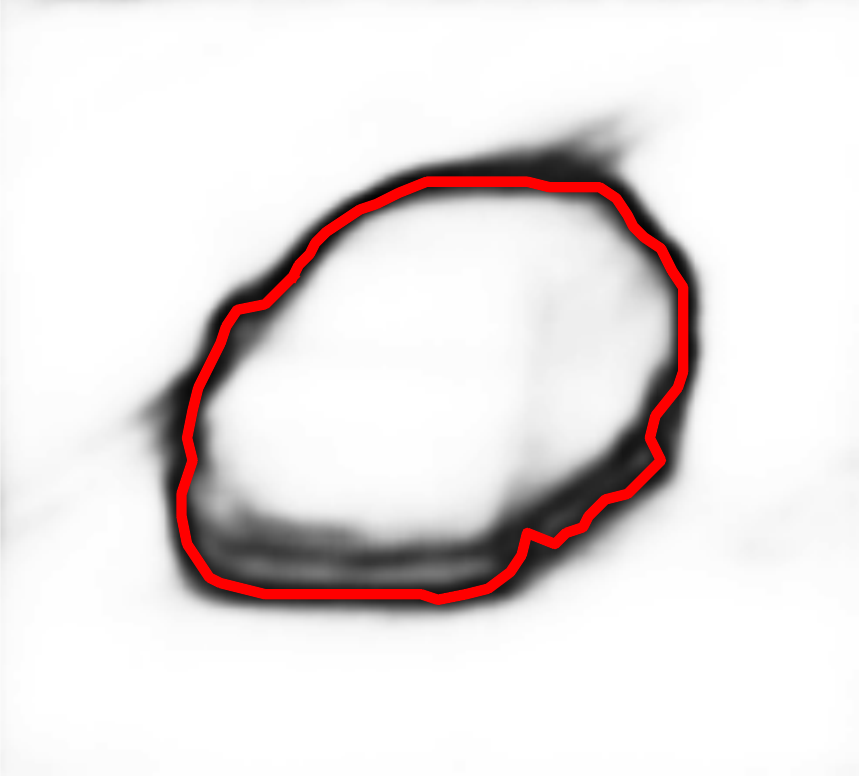}%
	\includegraphics[width=0.2\linewidth]{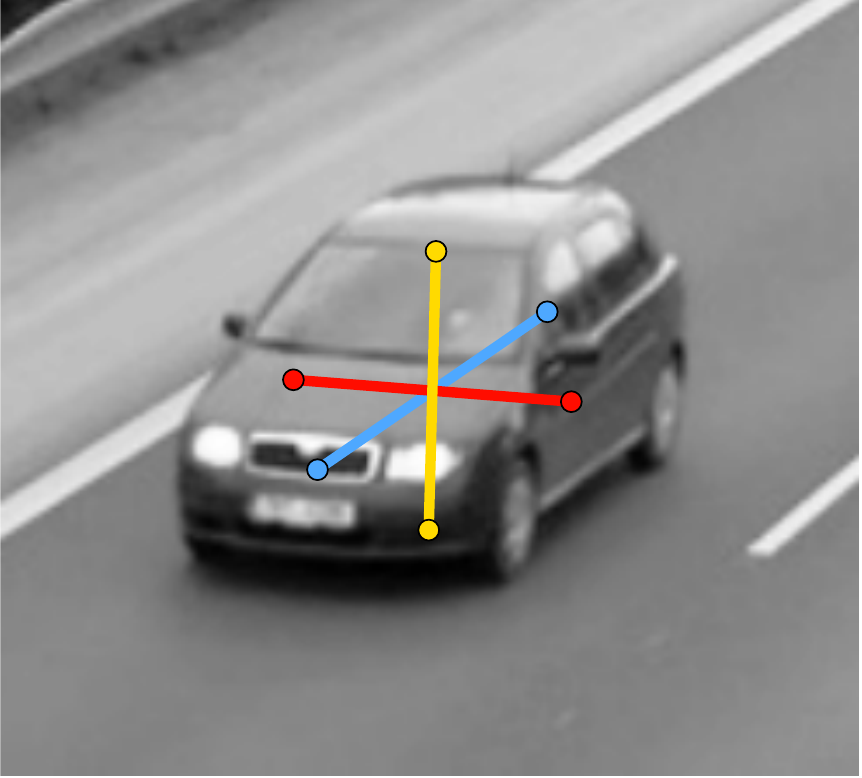}%
	\includegraphics[width=0.2\linewidth]{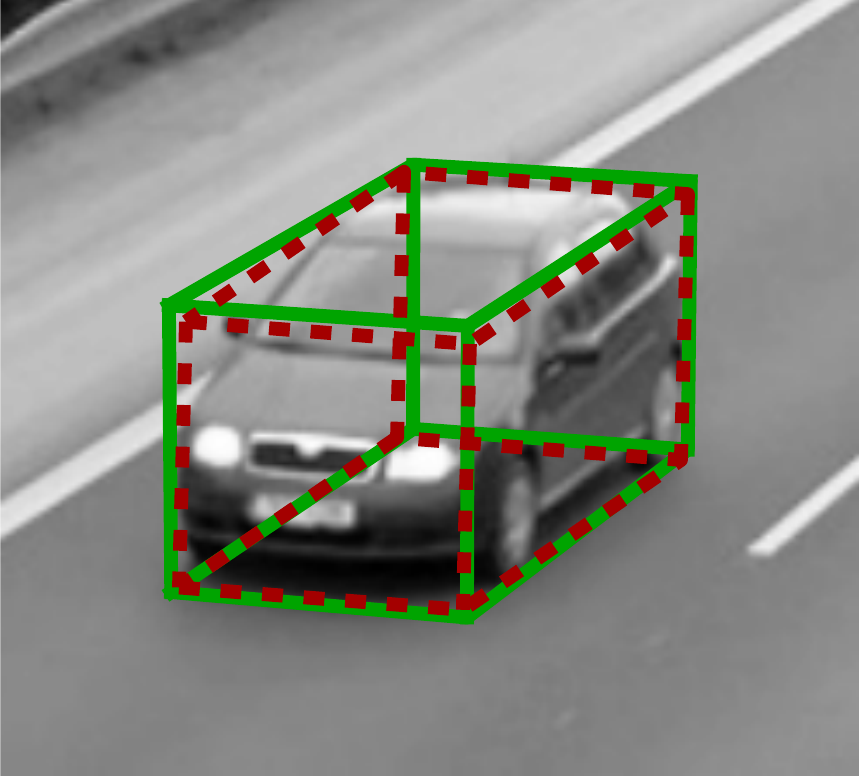}

	\caption{Estimation of 3D bounding box. \textbf{Left to right:} image with vehicle 2D bounding box, output of contour object detector \cite{Yang2016}, our constructed contour, estimated directions towards vanishing points, ground truth (\textbf{green}) and estimated (\textbf{red}) 3D bounding box.} \label{fig:3DBBEstimationTestTime}
\end{figure}

As proposed by \cite{Dubska2014}, the vehicle's contour and vanishing points are required for the bounding box construction. Therefore, it is necessary to estimate the contour and vanishing points for the vehicle. 
For estimating the vehicle contour, we use Fully Convolutional Encoder-Decoder network designed by Yang et al. \cite{Yang2016} for general object contour detection and masks with probabilities of vehicles contours for each image pixel. To obtain the final contour, we search for global maxima along line segments from 2D bounding box centers to edge points of the 2D bounding box (see Figure \ref{fig:3DBBEstimationTestTime} for examples).

We found out that the exact position of the vanishing point is not required for 3D bounding box construction, but the directions to the vanishing points are much more important. Therefore, we use regression to obtain the directions towards the vanishing points and then assume that the vanishing points are in infinity. 

\begin{figure}
	\centering
	\includegraphics[width=\linewidth]{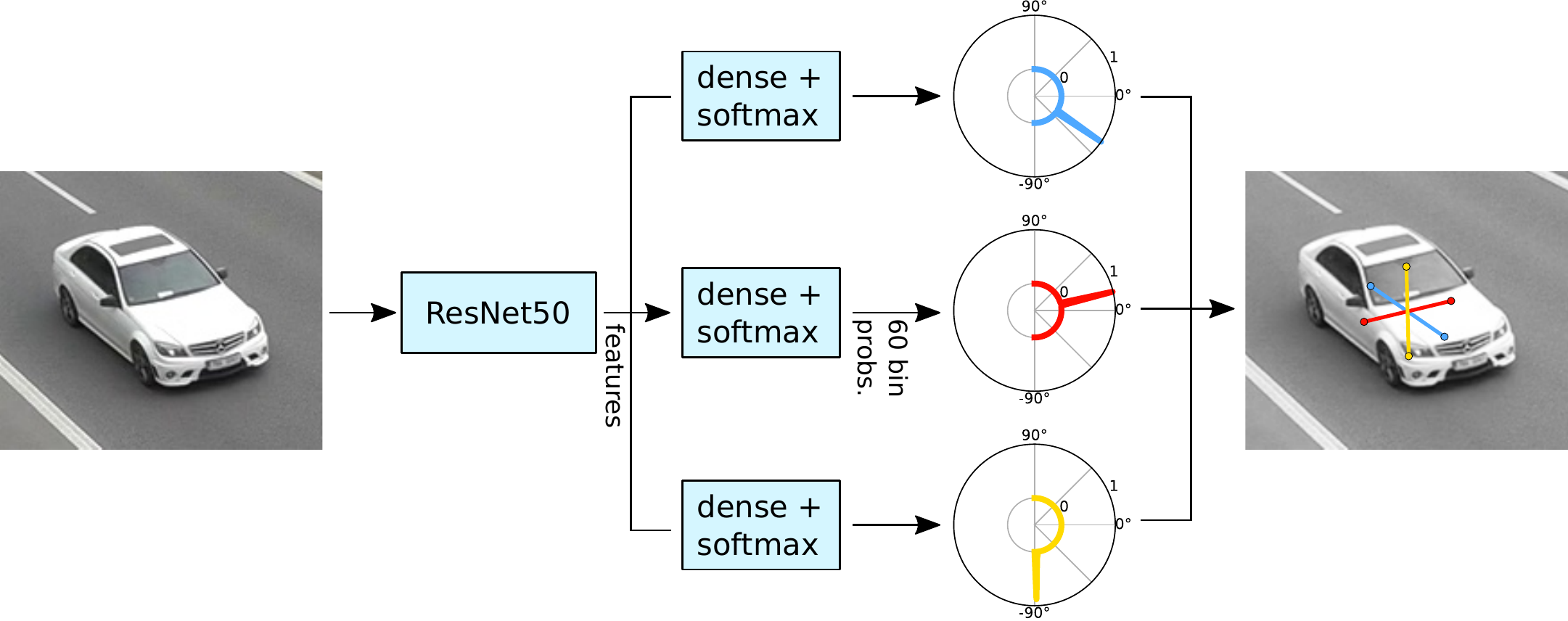}
	\caption{Used CNN for estimation of directions towards vanishing points. The vehicle image is fed to ResNet50 with 3 separate outputs which predict probabilities for directions of vanishing points as probabilities in a quantized angle space (60 bins from $-90^\circ$ to $90^\circ$). } \label{fig:VPRegreCNN}
\end{figure}

Following the work by Rothe et al. \cite{Rothe2016}, we formulated the regression of the direction towards the vanishing points as a classification task into bins corresponding to angles and we used ResNet50 \cite{He2015Resnet} with three classification outputs. We found this approach more robust than a direct regression.
We added three separate fully connected layers with softmax activation (one for each vanishing point) after the last average pooling in the ResNet50 (see Figure~\ref{fig:VPRegreCNN}). Each of these layers generates probabilities for each vanishing point belonging to the specific direction bin (represented as angles). We quantized the angle space by bins of $3^\circ$ from $-90^\circ$ to $90^\circ$ (60 bins per vanishing point in total).

As the training data for the regression we used BoxCars116k dataset (Section~\ref{sec:Dataset}) with the test samples omitted. The direction to vanishing points were obtained by method \cite{Dubska2014,Dubska2015ITS}; however, the quality of the ground truth bounding boxes was manually verified during annotation of the dataset and imprecise samples were removed by the annotators. 
To construct the lines on which the vanishing points are, we use the center of the 2D bounding box. Even though there is bias in the direction of the training data (some bins have very low number of samples), it is highly unlikely that for example, the first vanishing point direction will be close to horizontal.

With all this estimated information it is then possible to construct the 3D bounding box in both training and test time. It is important to note that by using this 3D bounding box estimation, it is possible to use this method outside the scope of traffic surveillance. It is only necessary to train the regressor of vanishing points directions. For the training of such a regressor, it is possible to use either the directions themselves or viewpoints on the vehicle and focal lengths of the images.  

Using this estimated bounding box, it is possible to unpack the vehicle image in test time without any additional information required. This enables the usage of the method when the traffic surveillance data are not available. The results in Section \ref{sec:ExperimentsEstimated3DBB} show that by using this estimated 3D bounding boxes, our method still significantly outperforms other convolutional neural networks without input modification.

\section{BoxCars116k Dataset} \label{sec:Dataset}

\begin{figure*}
	\centering
	\includegraphics[width=\linewidth]{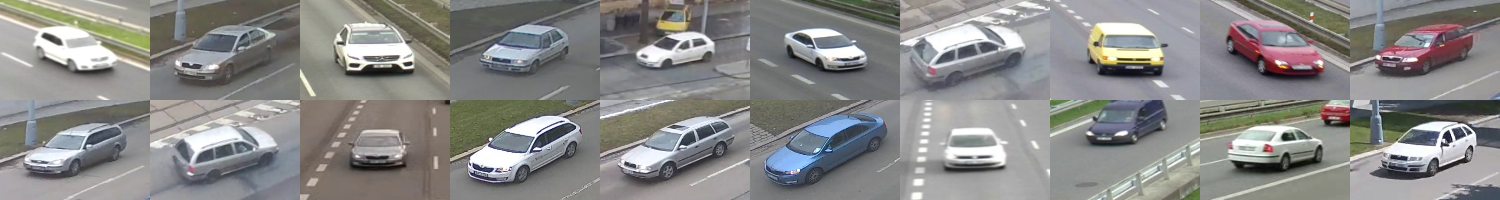} 
	\caption{Collate of random samples from the BoxCars116k dataset.} \label{fig:DatasetCollate}
\end{figure*}

We collected and annotated a new dataset \emph{BoxCars116k}. The dataset is focused on images taken from surveillance cameras as it is meant to be useful for traffic surveillance applications. We do not restrict that the vehicles are taken from the frontal side (Fig. \ref{fig:DatasetCollate}). We used surveillance cameras mounted near streets and tracked passing vehicles. The cameras were placed on various locations around Brno, Czech Republic and recorded the passing traffic from an arbitrary (reasonable) surveillance viewpoint. 
Each correctly detected vehicle (by Faster-RCNN \cite{Ren2015} trained on COD20k dataset \cite{Juranek2015}) is captured in multiple images, as it passes by the camera; therefore, we have more visual information about each vehicle.

\subsection{Dataset Acquisition}
The dataset is formed by two parts. The first part consists of data from \textit{BoxCars21k} dataset \cite{Sochor2016BoxCars} which were cleaned up and some imprecise annotations were then corrected (e.g. missing model years for some uncommon vehicle types). 

We also collected other data from videos relevant to our previous work \cite{Dubska2014,Dubska2015ITS,BrnoCompSpeed}. We detected all vehicles, tracked them and for each track collected images of the respective vehicle. We downsampled the framerate to $\sim12.5$ FPS to avoid collecting multiple and almost identical images of the same vehicle. 

The new dataset was annotated by multiple human annotators with an interest in vehicles and sufficient knowledge about vehicle types and models. The annotators were assigned to clean up the processed data from invalid detections and assign exact vehicle type (make, model, submodel, year) for each obtained track. While preparing the dataset for annotation, 3D bounding boxes were constructed for each detected vehicle using the method proposed by \cite{Dubska2014}. Invalid detections were then distinguished by the annotators based on these constructed 3D bounding boxes. In the cases when all 3D bounding boxes were not constructed precisely, the whole track was invalidated.

Vehicle type annotation reliability is guaranteed by providing multiple annotations for each valid track ($\sim 4$ annotations per vehicle). The annotation of a vehicle type is considered as correct in the case of at least three identical annotations. Uncertain cases were authoritatively annotated by the authors.

The tracks in \textit{BoxCars21k} dataset consist of exactly 3 images per track. In the new part of the dataset, we collect an arbitrary number of images per track (usually more than 3).
	
\subsection{Dataset Statistics}

\begin{figure}[t!]
	\centering
	\includegraphics[width=0.8\linewidth]{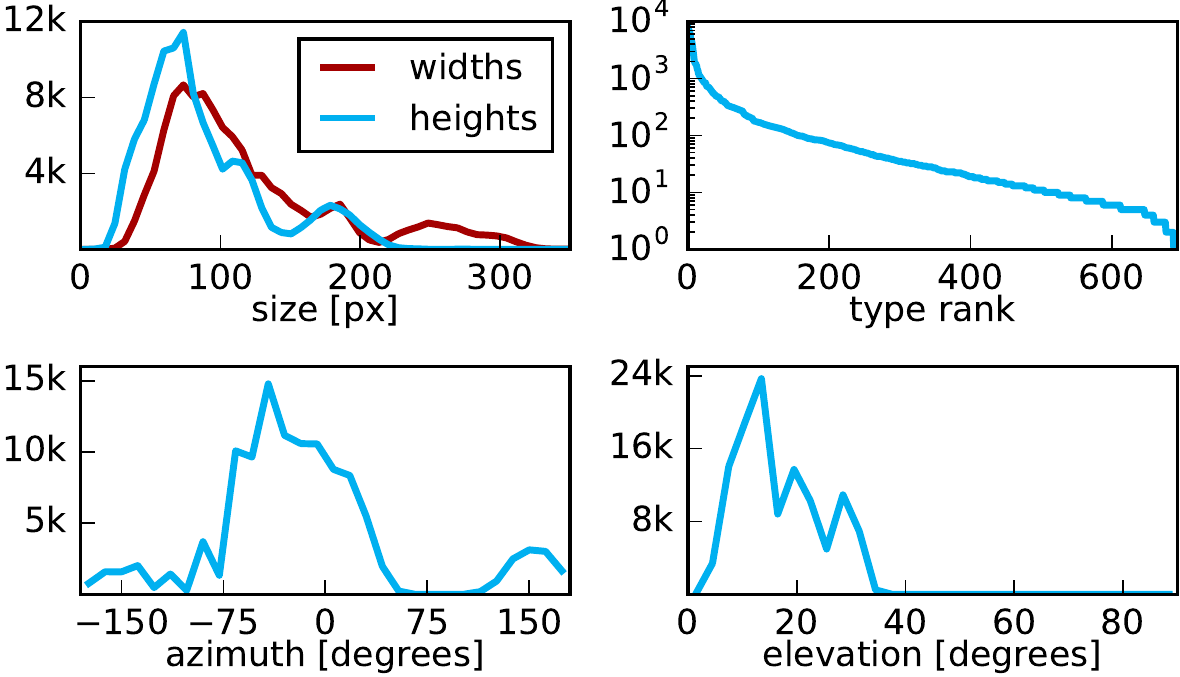}%
	\caption{BoxCars116k dataset statistics -- \textbf{top left:} 2D bounding box dimensions, \textbf{top right:} number of fine-grained types samples, \textbf{bottom left:} azimuth distribution ($0^\circ$ denotes frontal viewpoint), \textbf{bottom right:} elevation distribution.} \label{fig:DatasetStatistics}
\end{figure}

The dataset contains 27\,496
vehicles (116\,286
images) of 45 different makes with 693 fine-grained classes (make \& model \& submodel \& model year) collected from 137 different cameras with a large variation of viewpoints. Detailed statistics about the dataset can be found in  Figure~\ref{fig:DatasetStatistics} and the supplementary material. The distribution of types in the dataset is shown in Figure~\ref{fig:DatasetStatistics} (\textbf{top right}) and samples from the dataset are in Figure~\ref{fig:DatasetCollate}. The dataset also includes information about the 3D bounding box \cite{Dubska2014} for each vehicle and an image with a foreground mask extracted by background subtraction \cite{Stauffer1999,Zivkovic2004}. The dataset has been made publicly available%
\footnote{\url{https://medusa.fit.vutbr.cz/traffic}} 
for future reference and evaluation. 

Compared to ``web-based'' datasets, the new \textit{BoxCars116k} dataset contains images of vehicles relevant to traffic surveillance which have specific viewpoints (high elevation), usually small images, etc. Compared to other fine-grained surveillance datasets, our dataset provides data with a high variation of viewpoints (see Figure~\ref{fig:DatasetStatistics} and 3D plots in the supplementary material).

\subsection{Training \& Test Splits}

Our task is to provide a dataset for fine-grained recognition in traffic surveillance without any viewpoint constraint. Therefore, we have constructed the splits for training and evaluation in a way which reflects the fact that it is not usually known beforehand from which viewpoints the vehicles will be seen by the surveillance camera.

Thus, for the construction of the splits, we randomly selected cameras and used all tracks from these ca\-me\-ras for training and vehicles from the rest of the cameras for testing. In this way, we are testing the classification algorithms on images of vehicles from previously unseen cameras (viewpoints). This splits selection process implies that some of the vehicles from the test set may be taken under slightly different viewpoints from the ones that are in the training set.

We constructed two splits. In the first one (\textbf{hard}), we are interested in recognizing the precise type, in\-clu\-ding the model year. In the other one (\textbf{medium}), we omit the difference in model years and all vehicles of the same subtype (and potentially different model years) are present in the same class. We selected only types which have at least 15 tracks in the training set and at least one track in the testing set. The hard split contains 107 fine-grained classes with 11\,653 tracks (51\,691 images) for training and 11\,125 tracks (39\,149 images) for testing. Detailed split statistics can be found in the supplementary material.

\section{Experiments} \label{sec:Experiments}

We thoroughly evaluated our proposed algorithm on the BoxCars116k dataset. First, we evaluated how these methods improved classification accuracy with different nets, compared them to the state of the art, and analyzed how using approximate 3D bounding boxes influence the achieved accuracy. Then, we searched for the main source of improvements, analyzed improvements of different modifications separately, and also evaluated the usability of features from the trained nets for the task of vehicle type identity verification. 

In order to show that our modifications improve the accuracy independently on the used nets, we use several of them:
\begin{itemize}
	\item  \textbf{AlexNet} \cite{Krizhevsky2012}
	\item \textbf{VGG16}, \textbf{VGG19} \cite{Simonyan2014}
	\item \textbf{ResNet50}, \textbf{ResNet101}, \textbf{ResNet152} \cite{He2015Resnet}
	\item CNNs with Compact Bilinear Pooling layer \cite{Gao2016} in combination with VGG nets denoted as \textbf{VGG16+CBL} and \textbf{VGG19+CBL}.
\end{itemize}

As there are several options how to use the proposed modifications of input data and add additional auxiliary data, we define several labels which we will use:
\begin{itemize}
	\item \textbf{ALL} -- All five proposed modifications (Unpack, Color, ImageDrop, View, Rast).
	\item \textbf{IMAGE} -- Modifications working only on the image level (Unpack, Color, ImageDrop).
	\item \textbf{CVPR16} -- Modifications as proposed in our previous CVPR paper \cite{Sochor2016BoxCars} (Unpack, View, Rast -- however, the View and Rast modifications differ from those ones used in this paper as the original modifications do not distinguish between the frontal and rear side of vehicles).
\end{itemize}

\subsection{Improvements for Different CNNs} \label{sec:CNNImprovements}
The first experiment which was done was evaluation how our modifications have improved classification accuracy for different CNNs. 

\begin{table}[t]
	\caption{Summary statistics of improvements by our proposed modifications for different CNNs. The improvements over baseline CNNs are reported as single sample accuracy/track accuracy in percentage points. We also present classification error reduction in the same format. The raw numbers can be found in the supplementary material.} \label{tab:NetImprovementsSummary}.
	\begin{tabular}{clrrrr}
		\toprule
		& \textbf{modif.} & \multicolumn{2}{c}{\textbf{improvement [pp]}} & \multicolumn{2}{c}{\textbf{error reduction [\%]}}\\
		&   & mean      & best         & mean        & best         \\
		\midrule
		\parbox[t]{2mm}{\multirow{3}{*}{\rotatebox[origin=c]{90}{\textbf{medium}}}} &
		ALL      & 7.49/6.29 & 11.84/10.99 & 26.83/34.50 & 36.71/50.32 \\
		&IMAGE    & 7.19/6.15 & 12.09/11.63 & 27.38/36.21 & 35.23/49.55 \\
		&CVPR16   & 2.99/3.18 & 5.22/5.65   & 10.86/17.71 & 19.76/32.25 \\
		\midrule
		\parbox[t]{2mm}{\multirow{3}{*}{\rotatebox[origin=c]{90}{\textbf{hard}}}} &
		ALL    & 7.00/5.83 & 11.14/10.85 & 25.59/33.52 & 33.40/48.76 \\
		&IMAGE  & 6.74/5.81 & 11.02/10.53 & 26.12/35.95 & 33.04/47.33 \\
		&CVPR16 & 2.12/2.44 & 3.56/3.92   & 7.93/14.57  & 12.68/24.10 \\
		\bottomrule
	\end{tabular} 
\end{table}

All the nets were fine-tuned from models pre-trained on ImageNet \cite{ILSVRC15} for approximately 15 epochs which was sufficient for the nets to converge. We used the same batch size  (except for Res\-Net151, where we had to use a smaller batch size because of GPU memory limitations), the same initial learning rate and learning rate decay and the same hyperparameters for every net (initial learning rate $2.5\cdot 10^{-3}$, weight decay $5 \cdot 10^{-4}$, quadratic learning rate decay, loss is averaged over 100 iterations). 
We also used standard data augmentation techniques as a horizontal flip and randomly moving bounding box \cite{Simonyan2014}.
As ResNets do not use fully connected layers, we only use \textbf{IMAGE} modifications for them.

For each net and modification we evaluate the accuracy improvement of the modification in percentage points and also evaluate the classification error reduction.

The summary results for both medium and hard splits are shown in Table~\ref{tab:NetImprovementsSummary} and the raw results are in the supplementary material. As we have correspondences between the samples in the dataset and know which samples are from the same track, we are able to use mean probability across track samples and merge the classification for the whole track. Therefore, we always report the results in the form of \textit{single sample accuracy/whole track accuracy}. As expected, the results for whole tracks are much better than for single samples. For the traffic surveillance scenario, we consider to be more important the whole track accuracy as it is rather common to have a full track of observations of the same vehicle.

There are several things which should be noted about the results. The most important one is that our modifications significantly improve classification accuracy (up to \textbf{+12 percentage points}) and reduce classification error (up to \textbf{50\,\% error reduction}). Another important fact is that our new modifications push the accuracy much further compared to the original method \cite{Sochor2016BoxCars}. 

The table also shows that the difference between \textbf{ALL} modifications and \textbf{IMAGE} modifications is negligible and therefore it is reasonable to only use the \textbf{IMAGE} modifications. This also results in CNNs which just use the \textbf{Unpack} modification during test time as the other image modifications (Color, ImageDrop) are used only during fine-tuning of CNNs.

Moreover, the evaluation shows that the results are almost identical for the hard and medium split; therefore, we will only report additional results on the hard split, as it is the main goal to distinguish also the model years. The names for the splits were chosen to be consistent with the original version of dataset \cite{Sochor2016BoxCars} and the small difference between medium and hard split accuracies is caused mainly by the size of the new dataset.

\begin{table}[t!]
	\centering
		\caption{Comparison of different vehicle fine-grained recognition methods. Accuracy is reported as single image accuracy/whole track accuracy. Processing speed was measured on a machine with GTX1080 and CUDNN. ${}^*$ FPS reported by authors.} \label{tab:SOTAComparison}
	\begin{tabular}{l r r }
		\toprule
		\textbf{method} & \textbf{accuracy [\%]} &  \textbf{speed [FPS]}\\
		\midrule
		AlexNet \cite{Krizhevsky2012}  & $66.65$/$77.75$ & $963$\\		
		VGG16 \cite{Simonyan2014}  & $77.26$/$86.71$ & $173$\\		
		VGG19 \cite{Simonyan2014} & $76.74$/$86.06$ & $146$\\	
		Resnet50 \cite{He2015Resnet} & $75.48$/$84.61$ & $155$\\
		Resnet101 \cite{He2015Resnet} & $76.46$/$85.31$ & $95$\\
		Resnet152 \cite{He2015Resnet} & $77.68$/$86.20$ & $66$\\
		\midrule
		BCNN (VGG-M) \cite{Lin2015Bilinear} & $64.83$/$72.22$ & $87^*$\\
		BCNN (VGG16) \cite{Lin2015Bilinear}  & $69.64$/$78.56$ & $10^*$\\
		CBL (VGG16) \cite{Gao2016} & $70.38$/$80.11$ & $165$\\		
		CBL (VGG19) \cite{Gao2016}  & $70.69$/$80.26$ & $141$\\			
		PCM (AlexNet) \cite{Simon2015}  & $63.24$/$73.94$ & $15$\\
		PCM (VGG19) \cite{Simon2015}  & $75.99$/$85.24$ & $4$\\
		\midrule
		AlexNet + ALL (ours) & $77.79$/$88.60$  & 580\\	
		VGG16 + ALL (ours) & $\mathbf{84.13}$/$\mathbf{92.27}$ & 154\\	
		VGG19 + ALL (ours) & $\mathbf{84.12}$/$92.00$ & 133\\	
		VGG16+CBL + ALL (ours) & $75.06$/$83.42$ & 146\\	
		VGG19+CBL + ALL (ours)& $75.62$/$83.76$ & 126\\
		Resnet50 + IMAGE (ours)& $82.27$/$90.79$ & 151\\
		Resnet101 + IMAGE (ours)& $83.41$/$91.59$ & 93\\
		Resnet152 + IMAGE (ours) & $83.74$/$91.71$ & 65\\	
		\bottomrule
	\end{tabular}
\end{table}

\subsection{Comparison with the State of the Art}

In order to examine the performance of our method, we also evaluated other state-of-the-art methods for fine-grained recognition. We used three different algorithms for general fine-grained recognition with a published code. We always first used the code to reproduce the results in respective papers to ensure that we are using the published work correctly. All of the methods use CNNs and the used net influences the accuracy; therefore, the results should be compared with respective base CNNs. 

It was impossible to evaluate methods focused only on fine-grained recognition of vehicles as they are usually limited to frontal/rear viewpoint or require 3D models of vehicles for all the types. In the following text we define labels for each evaluated state-of-the-art method and describe details for the method separately.

\textbf{BCNN.} Lin et al. \cite{Lin2015Bilinear} proposed to use \textbf{B}ilinear \textbf{CNN}. We used VGG-M and VGG16 networks in a symmetric setup (details in the original paper), and trained the nets for 30 epochs (the nets converged around the $20^\mathrm{th}$ epoch). We also used image flipping to augment the training set. 

\textbf{CBL.} We modified compatible nets with \textbf{C}ompact \textbf{B}i\textbf{L}inear Pooling proposed by \cite{Gao2016} which followed the work of \cite{Lin2015Bilinear} and reduced the number of output features of the bilinear layers. We used the Caffe implementation of the layer provided by the authors and used 8\,192 features. We trained the net using the same hyper-parameters, protocol, and data augmentation as described in Section~\ref{sec:CNNImprovements}.

\begin{table}[t!]
	\centering
	\caption{Comparison of classification accuracy (percent) on the hard split with standard nets without any modifications, IMAGE modifications using 3D bounding box from surveillance data, and IMAGE modifications using estimated 3D BB (Section \ref{sec:Methodology3DBBEst}).} 
	\label{tab:Estimated3DBBEval}
	\begin{tabular}{l r r r}
		\toprule
		\textbf{net} & \textbf{no modification} & \textbf{GT 3D BB} & \textbf{estimated 3D BB}\\
		\midrule
		AlexNet & $66.65$/$77.75$ & $77.67$/$88.28$  &$74.81$/$87.30$\\
		VGG16 & $77.26$/$86.71$ & $83.79$/$92.23$ &$80.60$/$90.59$\\
		VGG19 & $76.74$/$86.06$  & $83.91$/$92.17$ & $81.43$/$91.57$\\
		VGG16+CBL & $70.38$/$80.11$ & $75.04$/$83.16$ &$72.83$/$82.92$\\
		VGG19+CBL & $70.69$/$80.26$  & $75.47$/$83.56$ &$73.09$/$83.09$\\
		ResNet50 & $75.48$/$84.61$ & $82.27$/$90.79$ &$79.60$/$90.40$\\
		ResNet101 & $76.46$/$85.31$  &  $83.41$/$91.59$ & $80.20$/$90.42$\\
		ResNet152 & $77.68$/$86.20$  & $83.74$/$91.71$ & $80.87$/$90.93$\\		
		\midrule
	\end{tabular}
\end{table}
\begin{figure*}
	\centering
	\includegraphics[width=0.25\linewidth]{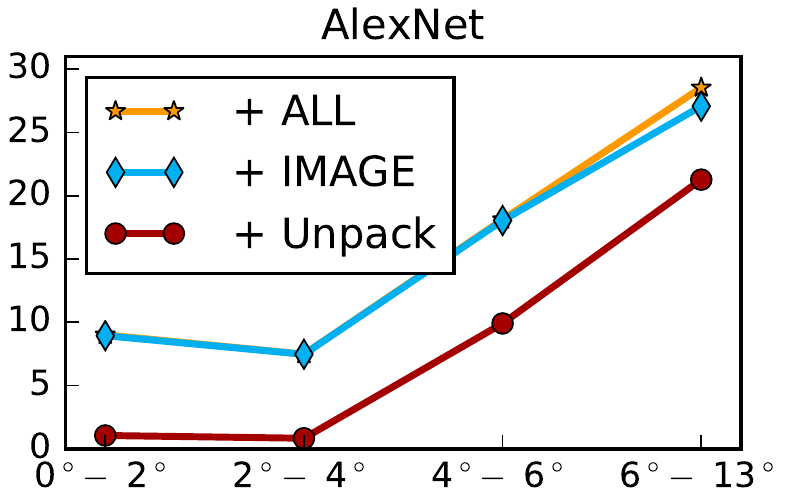}%
	\includegraphics[width=0.25\linewidth]{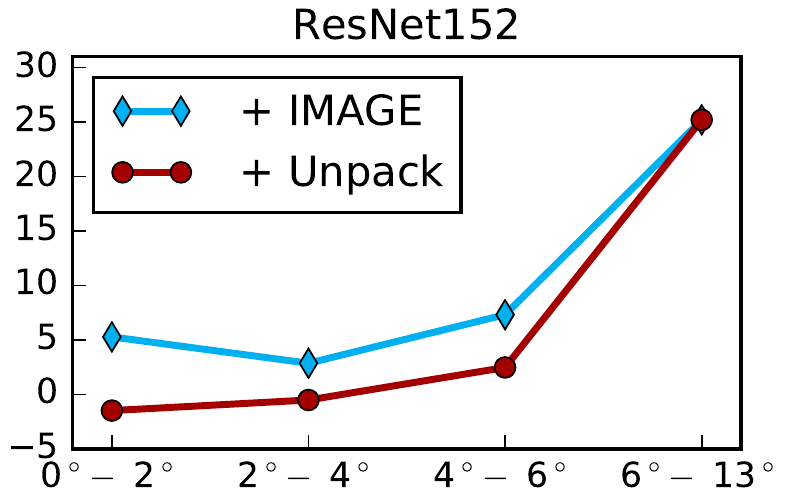}%
	\includegraphics[width=0.25\linewidth]{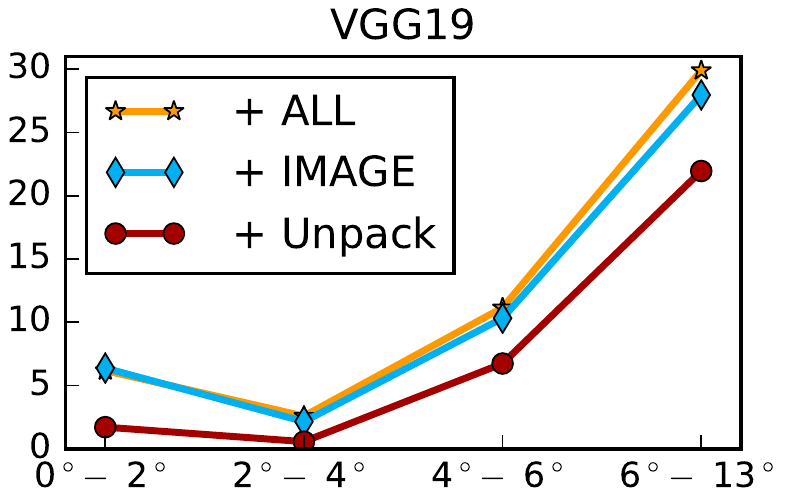}%
	\includegraphics[width=0.25\linewidth]{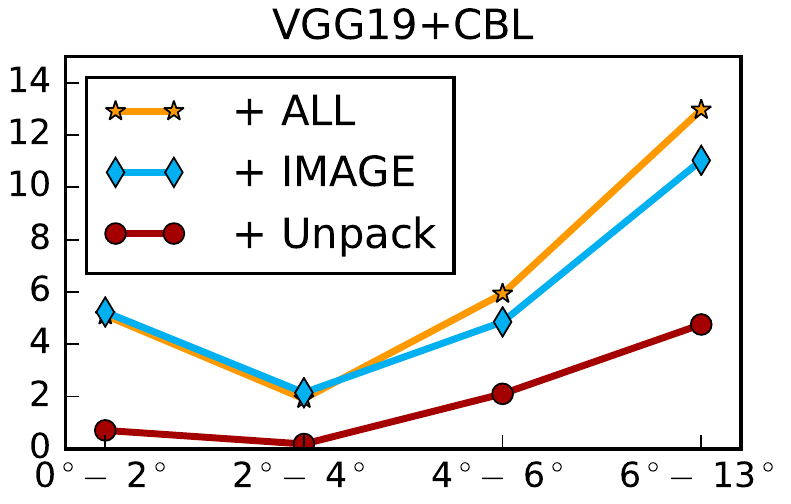}%
	\caption{Correlation of improvement relative to CNNs without modification with respect to train-test viewpoint difference. The $x$-axis contains bins viewpoint difference bins (in degrees), and the $y$-axis denotes improvement compared to base net in percent points, see Section \ref{sec:VPTErrorCorrelation} for details.  The graphs show that with increasing viewpoint difference, the accuracy improvement of our method increases. Only one representative of each CNN family (AlexNet, VGG, ResNet, VGG+CBL) is displayed -- results for all CNNs are in the supplementary material.} \label{fig:VptErrorCorrelation}
\end{figure*}

\textbf{PCM.} Simon et al. \cite{Simon2015} propose \textbf{P}art \textbf{C}on\-stellation \textbf{M}odels and use neural activations (see the paper for additional details) to get the parts of the model. We used AlexNet (BVLC Caffe reference version) and VGG19 as base nets for the method. We used the same hyper-parameters as the authors with the exception of fine-tuning number of iterations which was increased, and the $C$ parameter of used linear SVM was cross-validated on the training data. 

The results of all comparisons can be found in Table~\ref{tab:SOTAComparison}. As the table shows, our method significantly outperforms both standard CNNs \cite{Krizhevsky2012,Simonyan2014,He2015Resnet} and methods for fine-grained recognition \cite{Lin2015Bilinear,Simon2015,Gao2016}. The results for fine-grained recognition methods should be compared with the same used base network as for different networks, they provide different results. Our best accuracy ($84\,\%$) is better by a large margin compared to all other variants (both standard CNN and fine-grained methods).

In order to provide approximate information about the processing efficiency, we measured how many images different methods are able to process per second (referenced as FPS). The measurement was done with GTX1080 and CUDNN whenever possible. 
In the case of BCNN we reported the numbers as reported by the authors, as we were forced to save some intermediate data to disk because we were not able to fit all the data to memory ($\sim$200\,GB).
The results are also shown in Table~\ref{tab:SOTAComparison}; they show that our input modification decreased the processing speed; however, the speed penalty is small and the method is still usable for real-time processing.

\subsection{Influence of Using Estimated 3D Bounding Boxes instead of the Surveillance Ones} \label{sec:ExperimentsEstimated3DBB}
We also evaluated how the results will be influenced when, instead of using the 3D bounding boxes obtained from the surveillance data (long-time observation of video \cite{Dubska2014,Dubska2015ITS}), the estimated 3D bounding boxes (Section~\ref{sec:Methodology3DBBEst}) would be used instead.

The classification results are shown in Table~\ref{tab:Estimated3DBBEval}; they show that the proposed modifications still significantly improve the accuracy even if only the estimated 3D bounding box -- the less accurate one -- is used. 
This result is fairly important as it enables to transfer this method to different (non-surveillance) scenarios. The only additional data which is then required is a reliable training set of directions towards the vanishing points (or viewpoints and focal length) from the vehicles (or other rigid objects). 

\subsection{Impact of Training/Testing Viewpoint Difference} \label{sec:VPTErrorCorrelation}

\begin{figure}
	\centering
	\includegraphics[width=\linewidth]{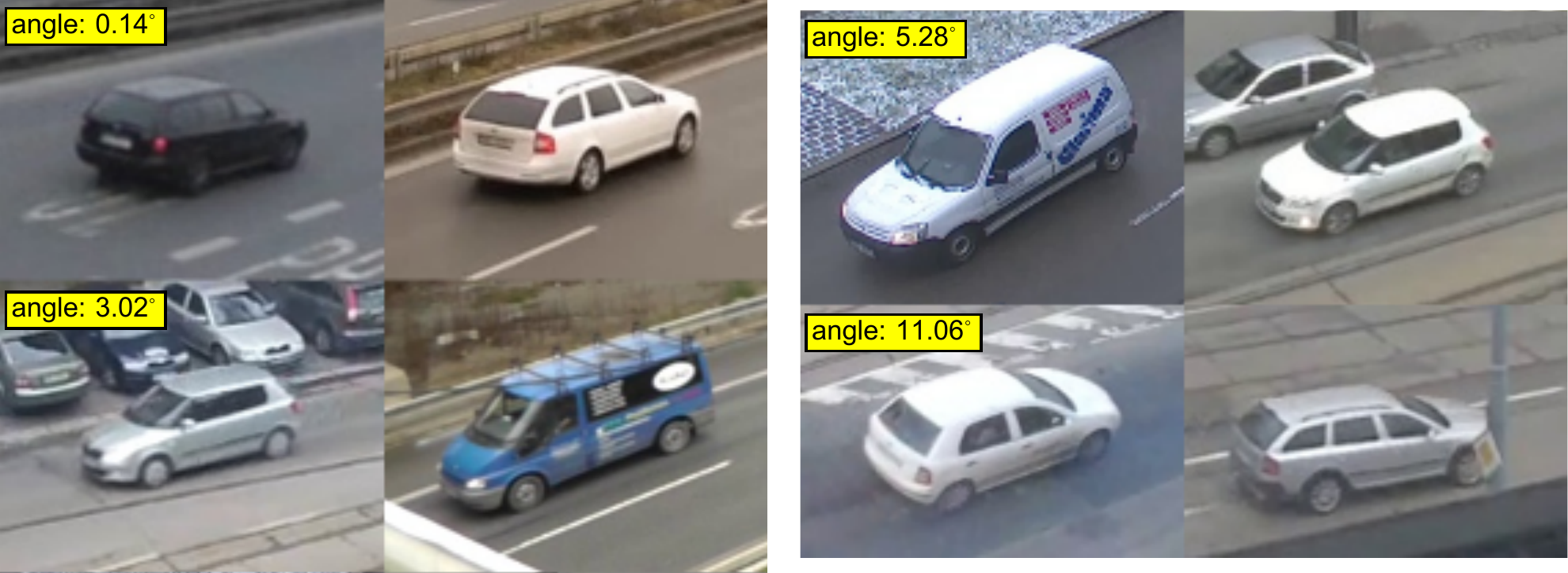}
	\caption{Examples of viewpoint difference between the training and testing sets.  Each pair shows a testing sample (left) and its corresponding ``nearest'' training sample (right); by ``nearest'' we mean the sample with the lowest angle between its viewpoint and the test sample's viewpoint.} 
	\label{fig:VptErrorCorrelationBins}
\end{figure}

We were also interested in finding out the main reason why the classification accuracy is improved. We have analyzed several possibilities and found out that the most important aspect is viewpoint difference. 

For every training and testing sample we computed the viewpoint (unit 3D vector from vehicles' 3D bounding boxes centers) and for each testing sample we found one training sample with the lowest viewpoint difference (see Figure~\ref{fig:VptErrorCorrelationBins}). Then, we divided the testing samples into several bins based on the difference angle. For each of these bins we computed the accuracy for the standard nets without any modifications and nets with the proposed modifications. There is 56\% of the test samples in the first bin ($0^\circ-2^\circ$), and in the middle bins there are 22\% and 17\% of test data. In the last bin, there are 5\% of the test data.
Finally, we obtained an improvement in percentage points for each modification and bin, by comparing the net's performance on the data in the bin with and without the modification harnessed. The results are displayed in Figure~\ref{fig:VptErrorCorrelation}.

There are several facts which should be noted. The first and most important is that the \textbf{Unpack} modification alone improves significantly the accuracy for larger viewpoint differences (the accuracy is improved by more than 20 percent points for the last bin). The other important fact, which should be noted, is that the other modifications (mainly \textbf{Color} and \textbf{ImageDrop}) improve the accuracy furthermore. This improvement is independent on the training-testing viewpoint difference.

\begin{figure*}
	\centering
	\includegraphics[width=0.25\linewidth]{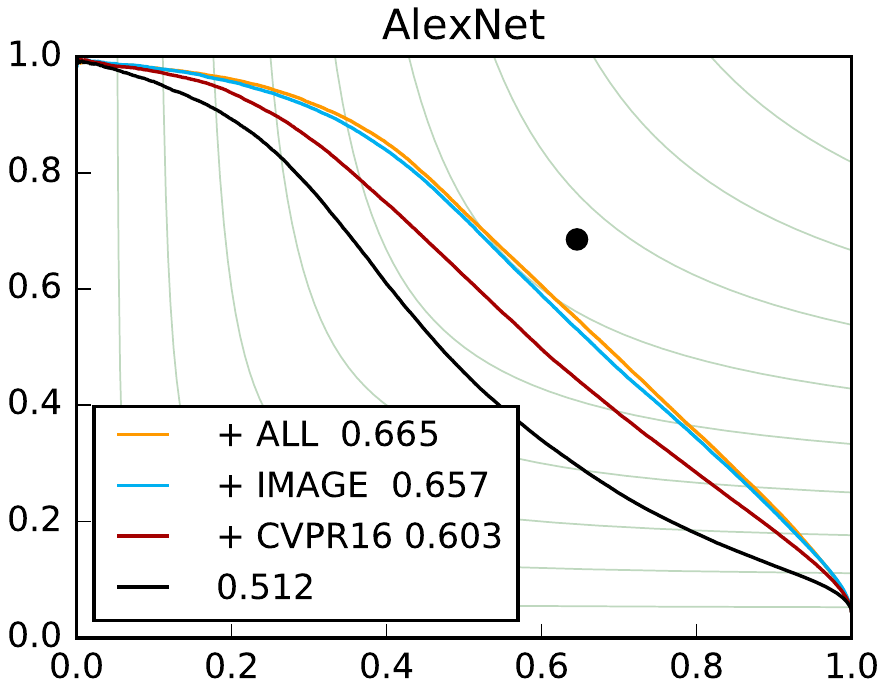}%
	\includegraphics[width=0.25\linewidth]{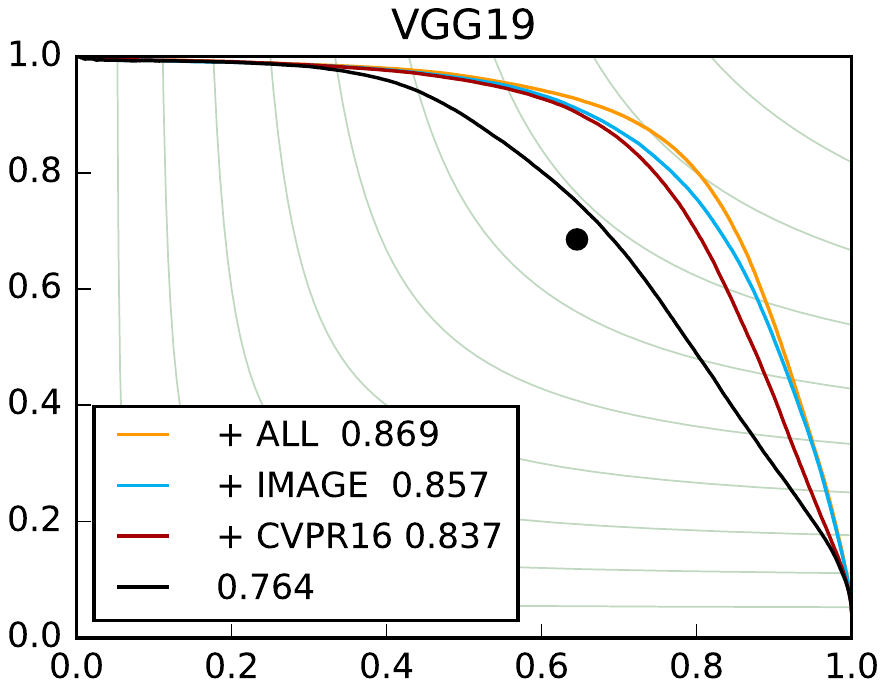}%
	\includegraphics[width=0.25\linewidth]{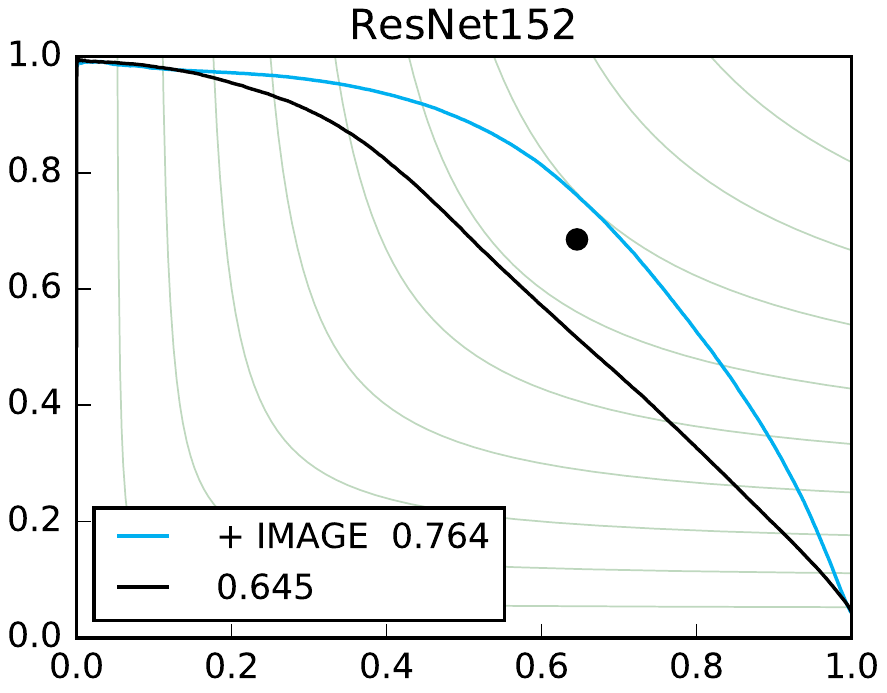}%
	\includegraphics[width=0.25\linewidth]{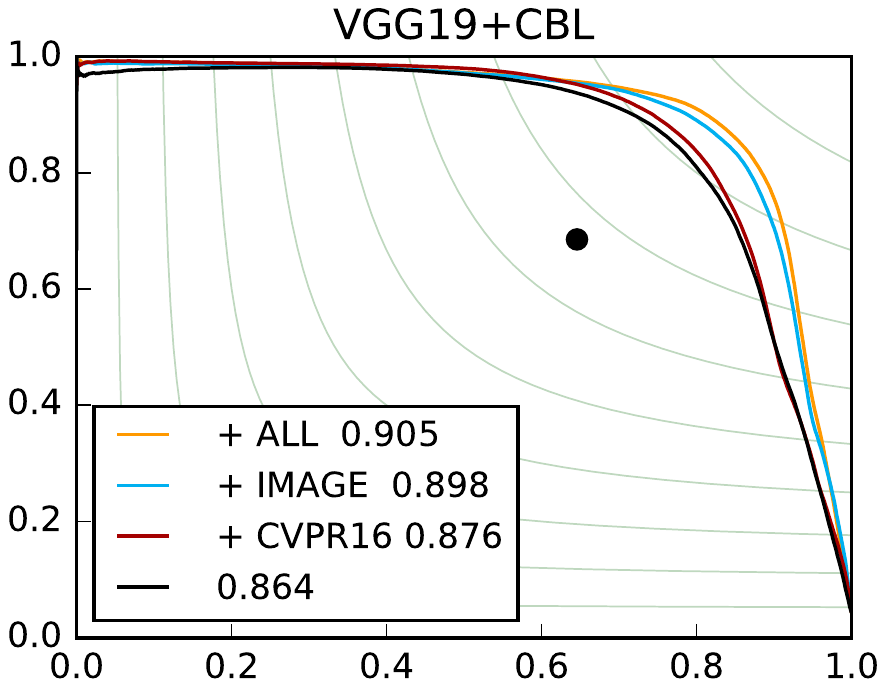}
	\caption{Precision-Recall curves for verification of fine-grained types. Black dots represent the human performance \cite{Sochor2016BoxCars}. Only one representative of each CNN family (AlexNet, VGG, ResNet, VGG+CBL) is displayed -- results for all CNNs are in the supplementary material. } \label{fig:Verification}
\end{figure*}

\subsection{Impact of Individual Modifications} 
\label{sec:SingleModificationImprovementExperiments}

We were also curious how different modifications by themselves help to improve the accuracy. 
We conducted two types of experiments which focus on different aspects of the modifications. The evaluation is not done on ResNets, as we only use \textbf{IMAGE} level modifications with ResNets; thus, we cannot evaluate Rast and View modifications with ResNets.

\begin{table}[t!]
	\centering
		\caption{Summary of improvements for different nets and modifications computed as $[\text{\textit{base net}} + \text{\textit{modification}}] - [\text{\textit{base net}}]$. The raw data can be found in the supplementary material.}
	\label{tab:ModificationAnalysisAddingHard}
		\begin{tabular}{lrr}
			\toprule
			 & \textbf{mean}        & \textbf{best}        \\
			\midrule
			Unpack    &$+2.11$/$+2.55$ & $+3.47$/$+4.37$ \\
			View     & $-0.32$/$-0.35$ & $+0.19$/$+0.31$ \\
			Rast     &  $-0.03$/$-0.04$ & $+0.30$/$+0.72$ \\
			Color     & $+3.17$/$+2.03$ & $+4.80$/$+3.60$ \\
			ImageDrop &  $+0.70$/$+0.20$ & $+1.53$/$+0.96$ \\
			\bottomrule
		\end{tabular}%
\end{table}

The first experiment is focused on the influence of each modification by itself. Therefore, we compute the accuracy improvement (in accuracy percent points) for the modifications as 
$[\text{\textit{base net}} + \text{\textit{modification}}] - [\text{\textit{base net}}]$, where $[\ldots]$ stands for the accuracy of the classifier described by its contents.
The results are shown in Table~\ref{tab:ModificationAnalysisAddingHard}. As it can be seen in the table, the most contributing modifications are \textbf{Color}, \textbf{Unpack}, and \textbf{ImageDrop}. 

The second experiment evaluates how a given modification contributed to the accuracy improvement when all of the modifications are used. Thus, the improvement is computed as 
$[\text{\textit{base net}} + \text{\textit{all}}] - [\text{\textit{base net}} + \text{\textit{all}} - \text{\textit{modification}}]$.
See Table~\ref{tab:ModificationAnalysisRemovingHard} for the results, which confirm the previous findings and \textbf{Color}, \textbf{Unpack}, and \textbf{ImageDrop} are again the most positive modifications.

\subsection{Vehicle Type Verification}
\label{sec:VerificationEvaluation}

Lastly, we evaluated the quality of features extracted from the last layer of the convolutional nets for the verification task. Under the term \emph{verification}, we understand the task to determine whether a pair of vehicle tracks share the same fine-grained type or not. In agreement with previous works in the field \cite{Taigman2014}, we use cosine distance between the features for the ve\-ri\-fi\-ca\-tion. 

We collected 5 million random pairs of vehicle tracks from the test part of \textit{BoxCars116k} splits and evaluate the verification on these pairs. As we used tracks which can have a different number of vehicle images, we used 9 random pairs of images for each pair of tracks and then used median distance between these image pairs as the distance between the whole tracks.

Precision-Recall curves and Average Precisions are shown in Figure~\ref{fig:Verification}. As the results show, our modifications significantly improve the average precision for each CNN in the given task. Moreover, as the figure shows, the method outperforms human performance (black dots in Figure~\ref{fig:Verification}), as reported in the previous paper \cite{Sochor2016BoxCars}.

\begin{table}[t!]
	\centering
	\caption{Summary of improvements for different nets and modifications computed as $[\text{\textit{base net}} + \text{\textit{all}}] - [\text{\textit{base net}} + \text{\textit{all}} - \text{\textit{modification}}]$. The raw data can be found in the supplementary material.}  
	\label{tab:ModificationAnalysisRemovingHard}
	\begin{tabular}{lrrrr}
		\toprule
		&   \textbf{mean}        & \textbf{best}        \\
		\midrule
		Unpack    & $+3.41$/$+3.48$ & $+6.93$/$+7.60$ \\
		View     & $-0.14$/$-0.15$ & $+0.36$/$+0.18$ \\
		Rast     &  $-0.03$/$-0.08$ & $+0.30$/$+0.20$ \\
		Color     &  $+3.42$/$+2.43$ & $+6.34$/$+6.18$ \\
		ImageDrop &  $+1.32$/$+0.77$ & $+4.24$/$+3.54$ \\
		\bottomrule
	\end{tabular}
	
\end{table}

\section{Conclusion} 
\label{sec:Conclusion}

This article presents and sums up multiple algorithmic modifications suitable for CNN-based fine-grained recognition of vehicles.  Some of the modifications were originally proposed in a conference paper \cite{Sochor2016BoxCars}, while others are results of the ongoing research.  
We also propose a method for obtaining the 3D bounding boxes necessary for the image unpacking (which has the largest impact on performance improvement) without observing a surveillance video, but only working with the individual input image.  This considerably increases the application potential of the proposed methodology (and the performance for such estimated 3D boxes is only somewhat lower than when ``proper'' bounding boxes are used).
We focused on a thorough evaluation of the methods: we coupled them with multiple state-of-the-art CNN architectures \cite{Simonyan2014,He2015Resnet}, and measured the contribution/influence of individual modifications.

Our method significantly improves the classification accuracy (up to \textbf{+12 percentage points}) and reduces the classification error (up to \textbf{50\,\% error reduction}) compared to the base CNNs. Also, our method outperforms other state-of-the-art methods \cite{Lin2015Bilinear,Simon2015,Gao2016} by \textbf{9 percentage points} in single image accuracy and by \textbf{7 percentage points} in whole track accuracy.

We collected, processed, and annotated a dataset \emph{BoxCars116k} targeted to fine-grained recognition of vehicles in the surveillance domain.  Contrary to a majority of existing vehicle recognition datasets, the viewpoints are greatly varying and correspond to surveillance scenarios; the existing datasets are mostly collected from web images and the vehicles are typically captured from eye-level positions.  This dataset has been made publicly available for future research and evaluation.

\section*{Acknowledgment}
This work was supported by The Ministry of Education, Youth and Sports of the Czech Republic from the National Programme of Sustainability (NPU II); project IT4Innovations excellence in science -- LQ1602.

{\small
	\bibliographystyle{IEEEtran}
	\bibliography{2017-ITS-BoxCars-bibliography}
}


\begin{IEEEbiography}[{\includegraphics[width=1in,height=1.25in,keepaspectratio]{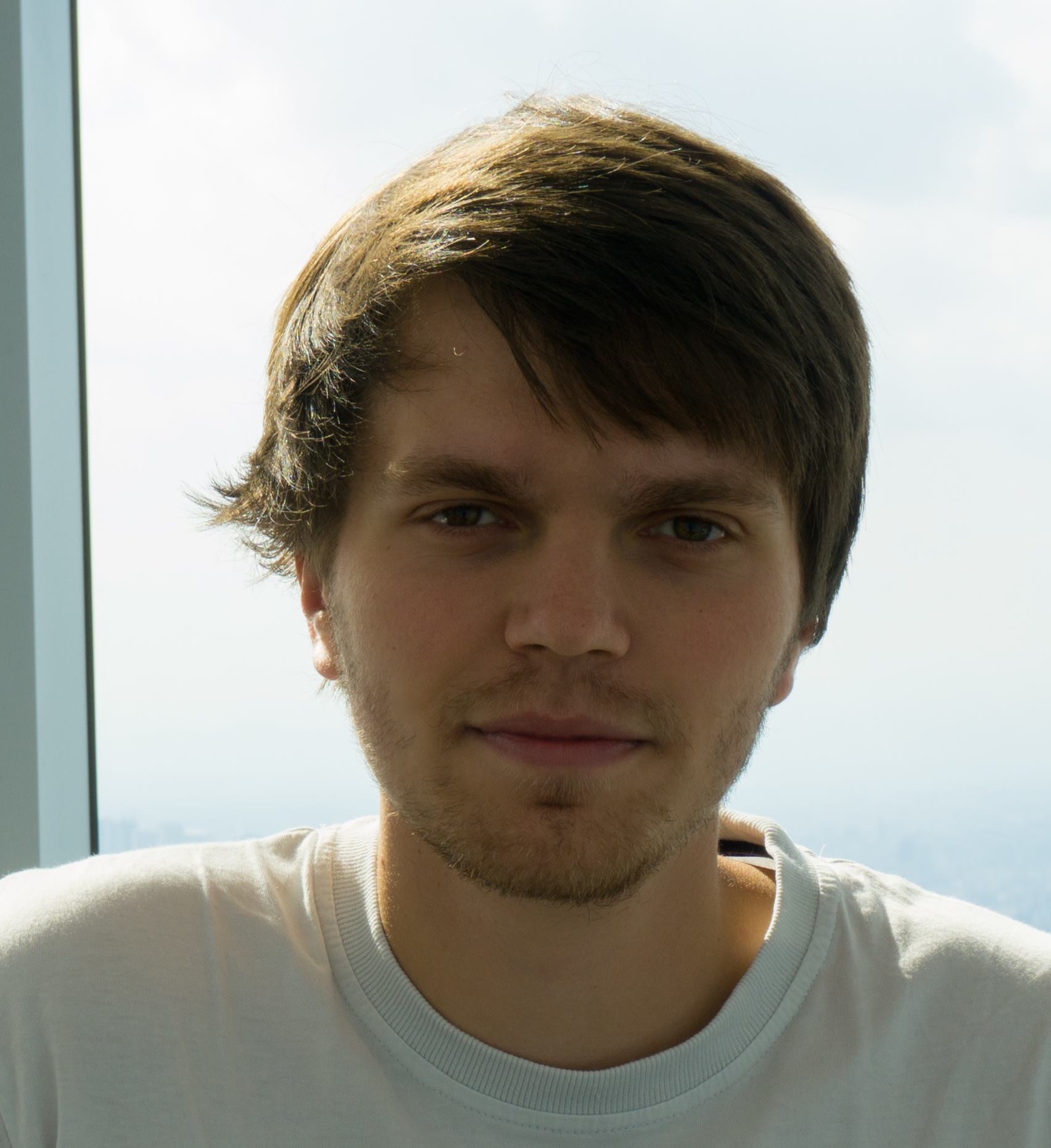}}]{Jakub Sochor}
received the M.S. degree from
Brno University of Technology (BUT), Brno, Czech
Republic. He is currently working toward the Ph.D.
degree in the Department of Computer Graphics
and Multimedia, Faculty of Information Technology,
BUT.
His research focuses on computer vision, particularly
traffic surveillance -- fine-grained recognition of vehicles and automatic speed measurement.
\end{IEEEbiography}
\vspace*{-1cm}
\begin{IEEEbiography}[{\includegraphics[width=1in,height=1.25in,clip,keepaspectratio]{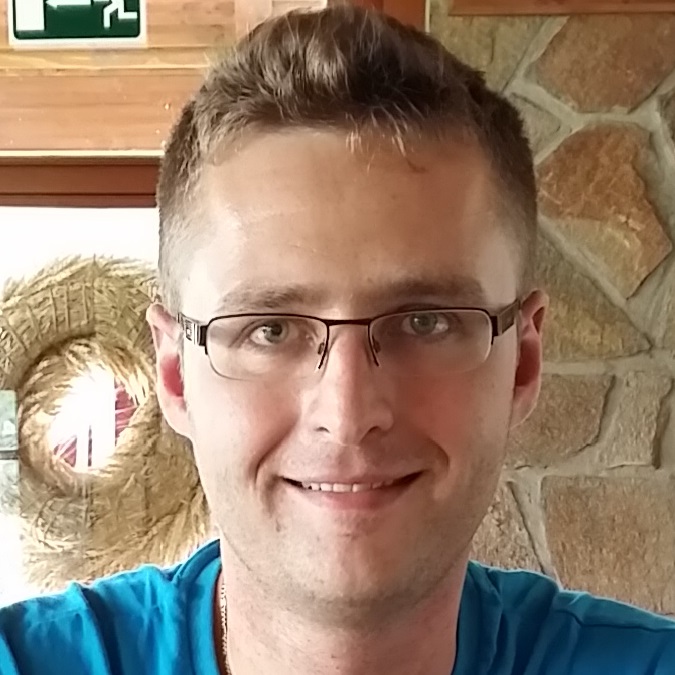}}]{Jakub Špaňhel}
received his M.S. degree from Faculty of Information Technology,
Brno University of Technology (BUT), Czech
Republic. He is currently working toward the Ph.D.
degree in the Department of Computer Graphics
and Multimedia, Faculty of Information Technology,
BUT.
His research focuses on computer vision, particularly
traffic analysis -- detection and re-identification of vehicles from surveillance cameras.
\end{IEEEbiography}
\vspace*{-1cm}
\begin{IEEEbiography}[{\includegraphics[width=1in,height=1.25in,clip,keepaspectratio]{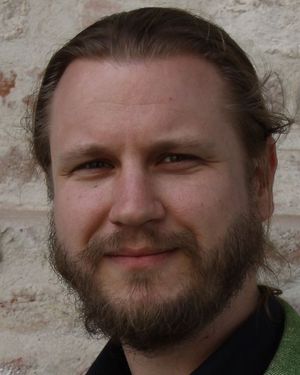}}]{Adam Herout}
	received his PhD from Faculty of Information Technology, Brno University of Technology, Czech Republic, where he works as a full professor and leads the Graph@FIT research group.  His research interests include fast algorithms and hardware acceleration in computer vision, with his focus on automatic traffic surveillance.
\end{IEEEbiography}


\vfill


\end{document}